\pdfoutput=1
\documentclass[entropy,article,accept,moreauthors,pdftex,10pt,a4paper]{mdpi} 

\firstpage{1} 
\articlenumber{}
\doinum{}
\pubvolume{xx}
\pubyear{2016}
\copyrightyear{2016}
\externaleditor{}
\history{Uploaded: \today}

\usepackage[final]{pdfpages}

\usepackage[utf8]{inputenc}

\usepackage{pdflscape}
\usepackage{booktabs}
\usepackage{amsmath}
\usepackage{mathtools}
\usepackage[utf8]{inputenc}
\usepackage{color}
\usepackage{multirow}
\usepackage{hyperref}
\usepackage{algorithm}
\usepackage[noend]{algpseudocode}
\floatname{algorithm}{Procedure}

\usepackage{caption}
\usepackage{subcaption}
\usepackage{adjustbox}
\usepackage{amsfonts}
\usepackage{wrapfig}

\newcommand{\ie}{\textit{i.e.} }
\newcommand{\rakel}{RA$k$EL } 
\newcommand{\rakeld}{RA$k$EL$d$ } 
\newcommand{\rakeldstop}{RA$k$EL$d$.} 
\newcommand{\rakelo}{RA$k$EL$o$ } 

\Title{How is a data-driven approach better than random choice in label space division for multi-label classification?}

\Author{Piotr Szymański $^{1,2}$*, Tomasz Kajdanowicz $^{1}$, Kristian Kersting $^{3}$}

\address{%
$^{1}$  \quad Department of Computational Intelligence, Wrocław University of Technology, Wybrzeże Stanisława Wyspiańskiego 27, 50-370 Wrocław, Poland\\
$^{2}$  \quad illimites Foundation,  Gajowicka 64 lok. 1, Wrocław, Poland\\
$^{3}$  \quad Computer Science Department, TU Dortmund University, August-Schmidt-Straße 4, 44221 Dortmund, Germany}

\corres{Correspondence: piotr.szymanski@pwr.edu.pl}

\abstract{We propose using five data-driven community detection approaches from social networks to partition the label space for the task of multi-label classification as an alternative to random partitioning into equal subsets as performed by RA$k$EL$d$: modularity-maximizing fastgreedy and leading eigenvector, infomap, walktrap and label propagation algorithms. We construct a label co-occurence graph (both weighted an unweighted versions) based on training data and perform community detection to partition the label set. We include Binary Relevance and Label Powerset classification methods for comparison. We use gini-index based Decision Trees as the base classifier. We compare educated approaches to label space divisions against random baselines on 12 benchmark data sets over five evaluation measures. We show that in almost all cases seven educated guess approaches are more likely to outperform RA$k$EL$d$ than otherwise in all measures, but Hamming Loss. We show that fastgreedy and walktrap community detection methods on weighted label co-occurence graphs are 85-92\% more likely to yield better F1 scores than random partitioning. Infomap on the unweighted label co-occurence graphs is on average 90\% of the times better than random paritioning in terms of Subset Accuracy and 89\% when it comes to Jaccard similarity. Weighted fastgreedy is better on average than RA$k$EL$d$ when it comes to Hamming Loss.}


\begin{document}


\section{Introduction}

Shanon's work on the unpredictability of information content inspired a search for the area of multi-label classification that requires more insight - where has the field still been using random approaches to handling data uncertainty when non-random methods could shed the light and provide the ability to make better predictions?

Interestingly enough, random methods are prevalent in well-cited, and multi-label classification approaches, especially in the problem of label space partitioning, which is a core issue in the problem-transformation approach to multi-label classification.

A great family of multi-label classification methods, called \textit{problem transformation} approaches, depends on converting an instance of a multi-label classification problem to one or more single-label single-class or multi-class classification problems, perform such classification and then convert the results back to multi-label classification results.

Such a situation stems from the fact that historically, the field of classification started out with solving single-label classification problems. In general a classification problem of understanding the relationship (function) between a set of objects and a set of categories that should be assigned to it. If the object is allocated to at most one category, the problem is called a single-label classification. When multiple assignments per instance are allowed, we are dealing with a multi-label classification scenario. 

In the single label scenario, in one variant we deal with a case when there is only one category, i.e. the problem is a binary choice - whether to assign a category or not, such a scenario is called single-class classification, ex. the case of classifying whether there is a car in the picture or not. The other case is when we have to choose at most one from many possible classes - such a case is called multi-class classification - i.e. classifying a picture with the dominant brand of cars present in it. The multi-label variant of this example would concern classifying a picture with all car brands present in it.

As both single- and multi-class classification problems have been considerably researched during the last decades. One can naturally see, that it is reasonable to transform the multi-label classification case, by dividing the label space, to a single- or a multi-class scenario. A great introduction to the field was written by \cite{tsoumakas_multi-label_2007}.

The two basic approaches to such transformation are Binary Relevance and Laber Powerset. Binary Relevance takes an a priori assumption that the label space is entirely separable, thus converting the multi-label problem to a family of single-class problems, one for every label - and making a decision - whether to apply it or not. Converting the results back to multi-label classification is based just on taking a union of all assigned labels. Regarding our example - binary relevance assumes that correlations between car brands are not important enough and discards them a priori by classifying with each brand separately. 

Label Powerset makes an opposite a priori assumption - the label space is non-divisible label-wise, and each possible label combination becomes a unique class. Such an approach yields a multi-class classification problem on a set of classes equal to the power set of the label set, i.e. growing exponentially if one treats all label combinations as possible. In practice, this would be intractable. Thus, as \cite{dembczynski_label_2012} note, Label Powerset is most commonly implement to handle only these combinations of classes that occur in the training set - and as such is prone to both overfitting. It is also - per \cite{tsoumakas_random_2011} - prone to differences in label combination distributions between the training set and the test set as well as to an imbalance in how label combinations are distributed in general. 

To remedy the overfitting of Label Powerset, Tsoumakas et. al. \cite{tsoumakas_random_2011} proposes to divide the label space into smaller subspaces and use Label Powerset in these subspaces. The source of proposed improvements come from the fact that it should be easier for Label Powerset to handle a large number of label space combinations in a smaller space. Two proposed approaches are called random k-labelsets (\rakel). \rakel comes in two variants - a label space partitioning \rakeld which divides the label set into $k$ disjoint subsets and \rakelo which is a sampling approach that allows overlapping of label subspaces. In our example, \rakel would randomly select a subset of brands and use the label powerset approach for brand combinations in all of the subspaces.

While these methods were developed, we saw advances in other fields brought us more and more tools to explore relations between entities in data. Social and complex networks have been flourishing after most of the well-established methods were published. In this paper, we propose a data-driven approach for label space partitioning in multi-label classification. While we tackle the problem of classification, our goal is to spark a reflection on how data-driven approaches to machine learning using new methods from complex/social networks can improve established procedures. We show that this direction is worth pursuing, by comparing method-driven and data-driven approaches towards partitioning the label space.

Why should one rely on label space division at random? Should not a data-driven approach be better than random choice? Are methods that perform simplistic a priori assumptions truly worse than the random approach? What are the variances of result quality upon label space partitioning? Instead of selecting random subspaces of brands, we could consider that some city brands occur more often with each other, and less so with other suburban brands. Based on such a premise we could build a weighted graph depicting the frequency of how often two-brands occur together in photos. Then using well-established community detection methods on this graph, we could provide a data-driven partition for the label space.

In this paper, we wish to follow Shanon's ambition to search for a data driven solution, an approach of finding structure instead of accepting uncertainty. We run RA$k$EL$d$ on 12 benchmark data sets, with up different values of parameter $k$ - taking $k$ equal to 10\%, 20\%, $\dots$, 90\% of the label set size. We draw 250 distinct partitions of the label set into subsets of $k$ labels, per every value of $k$. In case there are less than 250 possible partitions of the label space (for ex. because there are less than 10 labels), we consider all possible partitions. We then compare these results against the performance of methods based both on a priori assumptions - Binary Relevance and Label Powerset - and also well-established community detection methods employed in social and complex network analysis to detect patterns in the label space. For each of the measures we state three hypotheses:

\begin{itemize}
\item[] \textbf{RQ1}: a data-driven approach performs statistically better than the average random baseline
\item[] \textbf{RQ2}: a data-driven approach are more likely to outperform \rakeld than methods based on a priori assumptions
\item[] \textbf{RQ3}: a data-driven approach has a higher likelihood to outperform \rakeld in the worst case than methods based on a priori assumptions
\item[] \textbf{RQ4}: a data-driven approach is more likely to perform better than \rakeld, than otherwise, \ie the worst-case likelihood is greater than $0.5$
\end{itemize}

We describe used methods in Section \ref{sec:methods} and compare the results of the likelihood of an data-driven approach being better than randomness in Section \ref{sec:results}. We provide technical detail of the experimental scenario in Section \ref{sec:experiments}. We conclude with main findings of the paper and future work in section \ref{sec:finish}.

\section{Related work}

Our study builds on two kinds of approaches to multi-label classification: problem transformation and ensemble. We extend the Label Powerset problem transformation method by employing an ensemble of classifiers to classify obtained partitions of the label space separately. We show that partitioning the label space randomly - as is done in the RAkEL approach - can be improved by using a variety of methods to infer the structure of partitions from the training. We extend the original evaluation of random $k$-labelsets performance using a larger sampling of the label space and providing deeper insight into how \rakeld performs. We also provide some insights into the nature of random label space partitioning in \rakeldstop Finally we provide alternatives to random label space partitioning that are very likely to yield better results than \rakeld depending on the selected measure, and we show which methods to use depending on the generalization strategy.

Classifier chains \cite{read_classifier_2011} approach to label space partitioning is based on a bayesian conditioning scheme, in which labels are ordered in a chain and then the n-th classification is performed taking into account the output of the last n-1 classifications. These methods suffer from a variety of challenges: the results are not stable when ordering of labels in the chain changes and finding the optimal chain is NP-hard. Existing methods that optimize towards best quality cannot handle more than 15 labels in a data set (ex. Bayes-Optimal PCC \cite{DBLP:conf/icml/DembczynskiCH10}). Also in every classifier chain approach one always needs to train at least the same number of classifiers as there are labels, and if ensemble approaches are applied - much more. In our approach we use community detection methods to divide the label space into a fewer number of cases to classify as multi-class problems, instead of transforming to a large number of single-class problems that are interdependent. We also do not strive to find the directly optimal solution to community detection problems on the label co-occurence graph, to avoid overfitting - instead we perform approximations of the optimal solutions. This approach provides a large overhead over random approaches. We note that it would be an interesting question whether the random orderings in classifier chains are as suboptimal of a solution, as random partitioning turns out to be in label space partitioning. Yet it is not a subject of the study and is open to further research.

Tsoumakas et. al.'s \cite{tsoumakas_effective_2008} HOMER is a method of 2-step hierarchical multi-label classification in which the label space is divided based on label assignment vectors and then observations are classified first with cluster metalabels, and finally for each cluster they were labeled with, they are classified with labels of that cluster. We do not compare to HOMER directly in this article, due to a different nature of the classification scheme, as the subject of this study is to evaluate how data-driven label space partititioning using complex/social network methods - which can be seen as weak classifiers (as all objects are assigned automatically to all subsets) - can improve the random partitioning multi-label classification. HOMER uses a strong classifier to decide which object should be classified in which subspace. Although we do not compare directly to HOMER, due to difference in the classification scheme and base classifier, our research shows similarities to Tsoumakas's result that abandoning random label space partitioning for k-means based educaed guess improves classification results. Thus our results are in accord, yet we provide a much wider study as we have performed a much larger sampling of the random space then the authors of HOMER in their method describing paper.

Madjarov et. al. \cite{madjarov_extensive_2012} compare performance of a 12 of multi-label classifiers on 11 benchmark sets evaluated by 16 measures. To provide statistical support they use a Friedman multiple comparison procedure with a post-hoc Nemenyi test. They include the \rakelo procedure in their study, \ie the random label subsetting instead of partitioning. They do not evaluate the partitioning strategy \rakeld - which is the main subject of this study. Our main contribution - the study of how \rakeld performs against more informed approaches, therefore fills the unexplored space of Madjarov's et. al.'s extensive comparison. Note that, due to computational limits, we use CART decision trees instead of SVM as single-label base classifier, as explained in \ref{sec:expdesign}.

Zhang et. al. \cite{zhang2014review} review theoretical aspects and reported experimental performance of 8 multi-label algorithms and categorize them by the order of correlations taken into account and evaluation measure that they try to optimize. We follow the idea of their review table in Table \ref{tab:finalrh} with entries related to data-driven approaches for label space partitioning under a flat classification scheme.

\section{Multi-label classification}
\label{sec:methods} 

In this section we aim to provide a more rigid description of methods we use in the experimental scenario. We start by formalizing the notion of classification. Classification aims to understand a phenomenon, a relationship (function $f: X \rightarrow Y$) between objects and categories, by generalizing from empirically collected data $D$:
\begin{itemize}
\item objects are represented as feature vectors $\bar{x}$ from the input space $X$
\item categories, i.e. labels or classes come from a set $L$ and it spans the output space $Y$:
\begin{itemize}

\item in case of single-label single-class classification $|L|=1$ and $Y = \big\{0, 1 \big\}$
\item in case of single-label multi-class classification $|L|>1$ and $Y = \big\{0, 1, \dots, |L|\big\}$ 
\item in case of multi-label classification $Y = 2^L$

\end{itemize}

\item the empirical evidence collected: $D = (D_{x}, D_{y}) \subset X \times Y$
\item a quality criterion function $q$
\end{itemize}

In practice the empirical evidence $D$ is split into two groups: the training set for learning the classifier and the test set to use for evaluating the quality of classifier performance. For the purpose of this section we will denote $D_{train}$ as the training set. 

The goal of classification is to learn a classifier $h: X \rightarrow Y$ such that $h$ generalizes $D_{train}$ in a way that maximizes $q$. 

We are focusing on problem-transformation approaches that perform multi-label classification by transforming it to a single-label classification and convert the results back to the multi-label case. In this paper we use CART decision tress as the single-label base classifier. CART decision trees are a single-label classification method capable of both single- and multi-class classification. A decision tree constructs a binary tree in which every node performs a split based on the value of a chosen feature $X_i$ from the feature space $X$. For every feature, a threshold is found that minimizes an impurity function calculated for a threshold on the data available in the current node's subtree. For the selected pair of a feature $X_i$ and a threshold $\theta$ a new split is performed in the current node. Objects with values of the feature lesser than $\theta$ are evaluated in the left subtree of the node, while the rest in the right. The process is repeated recursively for every new node in the binary tree until the depth limit selected for the method is reached or there is just one observation left to evaluate. In all our scenarios we use CART as the base single-label single- and multi-class base classifier.

Binary Relevance learns $|L|$ single-label single-class base classifiers $b_j : X \rightarrow \{0,1\}$ for each $L_j \in L$ and outputs the multi-label classification using a classifier $h(\bar{x}) = \{L_j \in L: b_{j}(\bar{x}) =1\}$.

Label Powerset constructs a bijection $lp: 2^{L} \rightarrow C$ between each subset of labels $L_i$ and a class $C_i$. Label Powerset then learns a single-label multi-class base classifier $b : X \rightarrow C$ and transforms its output to a multi-label classification result ${lp}^{-1}(b(\bar{x}))$. 

RA$k$EL$d$ performs a random partition of the label set $L$ into $k$ subsets $L_j|_{1}^{k}$. For each $L_j$ a Label Powerset classifier $b_j: X \rightarrow L_j$ is learned. For a given input vector $\bar{x}$ the RA$k$EL$d$ classifier $h$ performs multi-label classification with each $b_j$ classifier and then sums the results, which can be formally described as $h(\bar{x}) = \bigcup_{j=0}^{k} { b_j(\bar{x})}$. Following the RA$k$EL$d$ scenario from \cite{tsoumakas_random_2011} all partitions of the set $L$ into $k$ subsets are equally probable.

\section{The data-driven approach}

Having described the baseline random scenario of RA$k$EL$d$ we now turn to explaining how complex/social network community detection methods fit into a data-driven perspective for label space division. In this scenario we are transforming the problem exactly like RA$k$EL$d$, but instead of performing random space partitioning we construct a label co-occurence graph from the training data and perform community detection on this graph to obtain a label space division.

\subsection{Label co-occurrence graph}

We construct the label co-occurrence graph as follows. We start with and undirected co-occurrence graph $G$ with the label set $L$ as the vertex set and the set of edges constructed from all pairs of labels that were assigned together at least once to an input object $\bar{x}$ in the training set (here $l_{i,j,..}$ denote labels, \ie elements of the set $L$):

$$ E = \big\{ \{\lambda_i, \lambda_j\}: \big(\exists(\bar{x}, \Lambda) \in D_{train}\big) \big(\lambda_i \in \Lambda \wedge \lambda_j \in \Lambda\big)\big\} $$

One can also extend this unweighted graph $G$ to a weighted graph by defining a function $w: L \rightarrow \mathbb{N}$:

$$ w(\lambda_{i},\lambda_{j}) = \text{number of input objects } \bar{x} \text{ that have both labels assigned} =$$
$$=\Big|  \big\{ \bar{x}: (\bar{x}, \Lambda) \in D_{train}  \wedge \lambda_i \in \Lambda \wedge \lambda_j \in \Lambda\big\}\Big|$$

Using such graph $G$, weighted or unweighted, we find a label space division by using one of the following community detection methods to partition the graph's vertex set which is equal to the label set.

\subsection{Dividing the label space}

Community detection methods are based on different principles as different fields defined communnities differently. We are employing a variety of methods 

Modularity-based approaches such as the fast greedy \cite{clauset_finding_2004} and the spectral leading eigenvector algorithms are based on detecting a partition of label sets that maximizes the modularity measure by \cite{newman_finding_2003}. Behind this measure lies an assumption that \textit{true community structure in a network corresponds to a statistically surprising arrangement of edges} \cite{newman_finding_2003}, i.e. that a community structure in a real phenomenon should exhibit a structure different than an average case of a random graph, which is generated under a given \textit{null model}. A well-established null model is the configuration model, which joins vertices together at random, but maintains the degree distribution of vertices in the graph. 

For a given partition of the label set, the modularity measure is the difference between how many edges of the empirically observed graph have both ends inside of a given community, i.e. $e(C) = \{(u,v) \in E: u \in C \wedge v \in C \}$ versus how many edges starting in this community would end in a different one in the random case: $r(C) = \frac{\sum_{v \in C}{deg(v)}}{|E|} $. More formally this is $Q(C) = \sum_{c \in C} {e(c) - r(c)}$. In case of weights, instead of counting the number of edges the total weight of edges is used and instead of taking vertex degrees in $r$, the vertex strenghts are used - a precise description of weighted modularity can be found in Newman's paper \cite{newman2004analysis}.

Finding $\bar{C} = \text{argmax}_{C} {Q(C)}$ is NP-hard as shown by Brandes et. al. \cite{brandes_modularity-np-completeness_2006}. We thus employ three different approximation-based techniques instead: a greedy, a multi-level hierarchical and a spectral recursively dividing algorithm. 

The fast greedy approach works based greedy aggregation of communities, starting with singletons and merging the communities iteratively. In each iteration the algorithm merges two communities based on which merge achieves the highest contribution to modularity. The algorithm stops when there is no possible merge that would increase the value of the current partition's modularity. It's complexity is $O(N log2 N)$.

The leading eigenvector approximation method depends on calculating a modularity matrix for a split of the graph into two-communities. Such a matrix allows to rewrite the two-community definition of modularity in a matrix form which can be than maximized using the largest positive eigenvalue and signs of the corresponding elements in the eigenvector of the modularity matrix - negative ones assigned to one community, positive ones to another. The algorithm starts with all labels in one community and performs consecutive splits recursively until all elements of the eigenvector have the same sign, or the community is a singleton. The method is based on the simplest variant of spectral modularity approximation as proposed by Newman \cite{2006PhRvE}. It's complexity is $O(M + N^2)$.

Infomap algorithm concentrates on finding the community \textit{structure of the network with respect to flow and to exploit the inference-compression duality to do so} \cite{2009EPJST.178...13R}. It relies on finding a partition of the vertex set that minimizes the map equation. The map equation divides flows through the graph into intra-community ones and the between-community ones and takes into consideration and entropy-based \textit{frequency-weighted average length of codewords} used to label nodes in communities and inter-communities.  It's complexity is $O(M)$.

The label propagation algorithm \cite{2007PhRvE..76c6106R} assigns a unique tag to every vertex in the graph. Next it iteratively updates the tag of every vertex with the tag assigned to the majority of the elements neighbours. The updating order is randomly selected at each iteration. The algorithm stops when all vertices have tags that are in accord with the dominant tag in their neighbourhood.  It's complexity is $O(N + M)$.

The walktrap algorithm is based on the intuition that \textit{random walks on a graph tend to get “trapped” into densely connected parts corresponding to communities} \cite{2005physics12106P}. It starts with a set of singleton communities and agglomerates obtained communities in a greedy iterative approach based on how close two vertices are in terms of random-walk distance. More precisely each step merges two communities to maximize the decrease of the mean (averaged over vertices) of squared distances between a vertex and all the vertices that are in the vertex's community. The random walk distance between two nodes is measured as the $L^2$ distance between random walk probability distribution starting in each of the nodes. The distances are of the same maximum length provided as a parameter to the method. It's expected complexity is $O(N^2*log(N))$.

In complexity notations $N$ is the number of nodes and $M$ the number of edges.

\subsection{Classification scheme}

In our data-driven scheme, the training phase is performed as follows:
\begin{enumerate}
\item the label co-occurence graph is constructed based on the training data set
\item the selected community detection algorithm is executed on the label co-occurence graph
\item for every community $L_i$ a new training data set $D_i$ is created by taking the original input space with only the label columns that are present in $L_i$
\item for every community a classifier $h_i$ is learned on training set $D_i$
\end{enumerate}

The classification phase is performed by performing classification on all subspaces detected in the training phase and taking the union of assigned labels: $h(\bar{x}) = \bigcup_{j=0}^{k} { b_j(\bar{x})}$.


\section{Experiments and Materials}
\label{sec:experiments}

To prepare ground for results, in this section we describe which data sets we have selected for evaluation and why. Then we present and justify model selection decisions for our experimental scheme. Next we describe the configuration of our experimental environment. Finally we describe the measures used for evaluation.

\subsection{Data Sets}

Following Madjarov's study (\cite{madjarov_extensive_2012}) we have selected 12 different well-cited multi-label classification benchmark data sets. The basic statistics of datasets used in experiments, such as the number of data instances, the number of attributes, the number of labels, labels’ cardinality, density and the distinct number of label combinations are available online\cite{mulandata}. We selected the data sets to obtain a balanced representation of problems in terms of number of objects, number of labels and domains. At the moment of publishing this is one of the largest study of RA$k$ELd both in terms of data sets examined and in terms of random label partitioning sample count. This study also has exhibits higher ratio of number of data sets to number of methods that other studies.

The text domain is represented by 5 data sets: \textit{bibtex}, \textit{delicious}, \textit{enron}, \textit{medical}, \textit{tmc2007-500}. \textit{Bibtex} (\cite{katakis_multilabel_2008}) comes from the ECML/PKDD 2008 Discovery Challenge and is based on data from the Bibsonomy.org publication sharing and bookmarking website. It exemplifies the problem of assigning tags to publications represented as an input space of bibliographic metadata such as: authors, paper/book title, journal volume, etc. \textit{Delicious} (\cite{tsoumakas_effective_2008}) is another user-tagged data set. It spans over 983 labels obtained by scraping the 140 most popular tags from the del.icio.us bookmarking website, retrieving the 1000 most recent bookmarks, selecting the 200 most popular, deduplication and filtering tags that were used to tag less than 10 websites. For those labels, websites tagged with them were scraped and from their contents top 500 words ranked by $\chi^2$ method were selected as input features. \textit{Tmc2007} (\cite{tsoumakas_effective_2008}) contains an input space consisting of similarly selected top 500 words appearing in flight safety reports. The labels represent the problems being described in these reports. \textit{Enron} (\cite{klimt2004enron}) contains emails from senior Enron Corporation employees categorized into topics by the UC Berkeley Enron E-mail Analysis Project\footnote{\url{http://bailando.sims.berkeley.edu/enron_email.html}} with the input space being a bag of word representation of the e-mails. \textit{Medical} (\cite{read_classifier_2011}) data set is Medical Natural 
Language Processing Challenge\footnote{\url{http://www.computationalmedicine.org/challenge/}} challenge. The input space is a bag-of-words representation of patient symptom history and labels represent diseases following International Classification of Diseases. 

The multimedia domain consists of five datasets \textit{scene}, \textit{corel5k}, \textit{mediamill}, \textit{emotions} and \textit{birds}. The image data set \textit{scene} (\cite{boutell_learning_2004}) semantically indexes still scenes annotated with any of the following categories: beach, fall-foliage, field, mountain, sunset and urban. \textit{Birds} data set \cite{birds} represents a problem of matching bird voice recordings extracted features features with a the subset of 19 bird species that is present in the recording, each label represents one spiecies. This data set was introduced (\cite{6661934}) during the The 9th annual MLSP competition. A larger image set \textit{corel5k} (\cite{duygulu_object_2002}) containing normalized-cut segmented images clustered into 499 bins. The bins were labeled with subsets 374 labels. \textit{Mediamill} data set of annotated video segments (\cite{snoek_challenge_2006}) was introduced during the 2005 NIST TRECVID challenge\footnote{\url{http://www.science.uva.nl/research/mediamill/challenge/}}. It is annotated with 101 labels referring to elements observable on the video. The \textit{emotions} data set \cite{trohidis2008multi} represents the problem of automated detection of emotion in music, assigning a subset of music emotions based on the Tellegen-Watson-Clark model to each of the songs.

The biological domain is represented with two data sets: \textit{yeast} and \textit{genbase}. The \textit{yeast} \cite{elisseeff_kernel_2001} data set concerns the problem of assigning functional classes to genes of \textit{saccharomyces cerevisiae} genome. The \textit{genbase} \cite{Diplaris2005} data set represents the problem of assigning classes to proteins based on detected motifs that serve as input features.

\subsection{Experiment design}
\label{sec:expdesign}
Using 12 benchmark data sets evaluated with five performance measures we compare eight approaches to label space partitioning for multi-label classification:
\begin{itemize}
\item five methods that divide the label space based on structure inferred from the training data via label co-occurrence graphs - in both unweighted and weighted versions of the graphs
\item 2 methods that take an a priori assumption about the nature of the label space: Binary Relevance and Label Powerset
\item 1 random label space partitioning approach that draws partitions with equal probability - RA$k$EL$d$ 
\end{itemize}

In the random baseline (RA$k$EL$d$) we perform 250 samplings of random label space partitions into $k$-label subsets for each evaluated value of $k$. If a data set had more than 10 labels, we took values of $k$ ranging from 10\% to 90\% with a step of 10\%, rounding to closest integer number if necessary. In case of two data sets with a smaller number of labels i.e. \textit{scene} and \textit{emotions} we evaluated RA$k$EL$d$ for all possible label space partitions due to their low number. The number of label space division samples per dataset can be found in Table \ref{tab:samplings}.  As noone knows the true distribution of classification quality over label space partitions we've decided to use a large number of samples - 250 per each of the groups, 2500 alltogether - to get as close to a representative sample of the population, as was possible with our infrastructure limitations. 

As the base classifier, we use CART decision trees. While we recognize that the majority of studies prefer to use SVMs, we note that it is intractable to evaluate nearly 32500 samples of the random label space partitions using SVMs. We have thus decided to use a classifier that presents a reasonable trade-off between quality and computational speed.

We perform statistical evaluation of our approaches by comparing them to average performance of the random baseline of \rakeldstop We average \rakeld results per data set, which is justified by the fact that, this is the expected result one would get without performing extensive parameter optimization. Following Derrac et. al.'s \cite{Derrac20113} de facto standard modus operandi we use Friedman test with Iman-Davenport modifications to detect differences between methods and we check whether a given method is statistically better than the average random baseline using Rom's post-hoc pairwise test. We use these tests' results to confirm or reject \textbf{RH1}.

We do not perform statistical evaluation per group (\ie isolating each value of $k$ from 10\% to 90\%) due to lack of non-parametric repeated measure tests as noted by Demsar in the classic paper \cite{Demsar:2006}. 

Instead to account for variation, we consider the probability that a given data-driven approach to label space division is better than random partitioning. These probabilities were calculated per data set, as the fraction of random outputs that yielded worse results than a given method. Thus for example if infomap has 96.5\% probability of having higher better Subset Accuracy (SA) than the random approach in Corel5k, this means that on this data set infomap's SA score was better than scores achieved by 96.5\% of all RA$k$EL$d$ experiments. We check the median, the mean, and the minimal (\ie worst case) likelihoods. We use these results to confirm or reject \textbf{RH2}, \textbf{RH3} and \textbf{RH4}.

\subsection{Environment}

We used \texttt{scikit-multilearn} \cite{scikitmultilearn} - version 0.0.1- a scikit-learn API compatible library for multi-label classification in python that provides own implementation of several classifiers and uses \texttt{scikit-learn} \cite{scikit-learn} multi-class classification methods. All of the data sets come from \texttt{MULAN} \cite{mulan} data set library \cite{mulandata} and follow MULAN's division into the train and test subsets.

We use CART decision trees from the \texttt{scikit-learn} package - version 0.15 - with the Gini index as the impurity function. We employ community detection methods from the Python version of the \texttt{igraph} library \cite{igraph} for both weighted and unweighted graphs. Performance measures' implementation comes from the \texttt{scikit-learn} \texttt{metrics} package.

\subsection{Evaluation Methods}

Following Madjarov et. al.'s \cite{madjarov_extensive_2012} taxonomy of multi-label classification evaluation measures we use three example-based measures: hamming loss, subset accuracy and Jaccard similarity and a label-based measure - F1 as evaluated by two averaging schemes: micro and macro. In all following definitions:
\begin{itemize}
\item $X$ is the set of objects used in the testing scenario for evaluation
\item $L$ is the set of labels that spans the output space $Y$
\item $\bar{x}$ denotes an example object undergoing classification
\item $h(\bar{x})$ denotes the label set assigned to object $\bar{x}$ by the evaluated classifier $h$ 
\item $y$ denotes the set of true labels for the observation $\bar{x}$
\item $tp_j$, $fp_j$, $fn_j$, $tn_j$ are respectively true positives, false positives, false negatives and true negatives of the of label $L_j$, counted per label over the output of classifier $h$ on the set of testing objects $\bar{x} \in X$, \ie $h(X)$
\item the operator $[[p]]$ converts logical value to a number, i.e. it yields $1$ if $p$ is true and $0$ if $p$ is false
\end{itemize}

\subsubsection{Example-based evaluation methods}

Hamming Loss is a label-wise decomposable function counting the fraction of labels that were misclassified. $\otimes$ is the logical exclusive or. 

$$ \mathrm{HammingLoss}(h) = \frac{1}{|X|}\sum_{\bar{x} \in X}{\frac{1}{|L|} \sum_{L_{j} \in L}{[[ (L_{j} \in h(\bar{x})) \otimes (L_{j} \ \in y)  ]]} } $$

Accuracy score and subset 0/1 loss are instance-wise measures that count the fraction of input observations that have been classified exactly the same as in the golden truth. 

$$ \mathrm{SubsetAccuracy}(h) = \frac{1}{|X|}\sum_{\bar{x} \in X}{[[h(\bar{x}) = y ]] } $$

Jaccard similarity is a measure of the size of similarity between the prediction and the ground truth comparing what is the cardinality of an intersection of the two, compared to the union of the two. In other words what fraction of all labels taken into account by any of the prediction or ground truth were assigned to the observation in both of the cases.

$$ \mathrm{Jaccard}(h) = \frac{1}{|X|}\sum_{\bar{x} \in X}{ \frac{h(\bar{x}) \cap y}{h(\bar{x}) \cup y}  } $$

\subsubsection{Label-based evaluation methods}

The F1 measure is a harmonic mean of precision and recall where none of the two are more preferred than the other. Precision is the measure of how much the method is immune to the Type I error \ie falsely classifying negative cases as positives: \textit{false positives} or \textit{FP}. It is the fraction of correctly positively classified cases (\ie \textit{true positives}) to all positively classified cases. It can be interpreted as the probability that an object without a given label will not be labeled as having it. Recall is the measure of how much the method is immune to the Type II error \ie falsely classifying positive cases as negatives: \textit{false negatives} or \textit{FN}. It is the fraction of correctly positively classified cases (\ie \textit{true positives}) to all positively classified label. It can be interpreted as the probability that an object with a given label will be labeled as such.

These measures can be averaged from two perspectives that are not equivalent in practice due to a natural non-uniformity of distribution of labels among input objects in any testing set. Two averaging techniques are well-established as noted by \cite{Yang:1999:ESA:357367.357383}. 

Micro-averaging gives equal weight to every input object and performs a global agregation of true/false positives/negatives, averaging over all objects first. Thus:

$$\mathrm{precision}_{micro}(h) = \frac{\sum_{j=1}^{|L|}{{tp}_j}}{\sum_{j=1}^{|L|}{{tp}_{j} + {fp}_{j}}} $$

$$\mathrm{recall}_{micro}(h) = \frac{\sum_{j=1}^{|L|}{{tp}_j}}{\sum_{j=1}^{|L|}{{tp}_{j} + {fn}_{j}}} $$

$$\mathrm{F1}_{micro}(h) = 2 \cdot \frac{\mathrm{precision}_{micro}(h) \cdot \mathrm{recall}_{micro}(h)}{\mathrm{precision}_{micro}(h) + \mathrm{recall}_{micro}(h)} $$


In macro-averaging the measure is first calculated per label, then averaged over the number of labels. Macro averaging thus gives equal weight to each label, regardless of how often the label appears. 

$$\mathrm{precision}_{macro}(h, j) = \frac{{tp}_{j}}{{tp}_{j} + {fp}_{j}}$$
$$\mathrm{recall}_{macro}(h, j) = \frac{{tp}_{j}}{{tp}_{j} + {fn}_{j}}$$
$$\mathrm{F1}_{macro}(h, j) =  2 \cdot \frac{\mathrm{precision}_{macro}(h,j) \cdot \mathrm{recall}_{macro}(h,j)}{\mathrm{precision}_{macro}(h,j) + \mathrm{recall}_{macro}(h,j)} $$
$$\mathrm{F1}_{macro}(h) = \frac{1}{|L|} \sum_{j=1}^{|L|}{\mathrm{F1}_{macro}(h, j)} $$


\section{Results and Discussion}
\label{sec:results}

We describe performance per measure first and then look at how methods behave across measures. We evaluate each of the research hypotheses: \textbf{RH1}-\textbf{RH4} for each of the measures. We then look how these methods performed across data sets. We compare the median and the mean of achieved probabilities to assess average advantage over randomness, the higher the better. We compare the median and the means, as in some cases methods admit a single worst-performing outlier while in general providing large advantage over random approaches. We also check how each method's perform in the worst case i.e. what is the minimum probability of it being better than randomness in label space division? 

\subsection{Micro-averaged F1 score}
\begin{wrapfigure}{L}{0.33\textwidth}
  \begin{center}
    \includegraphics[width=.3\textwidth]{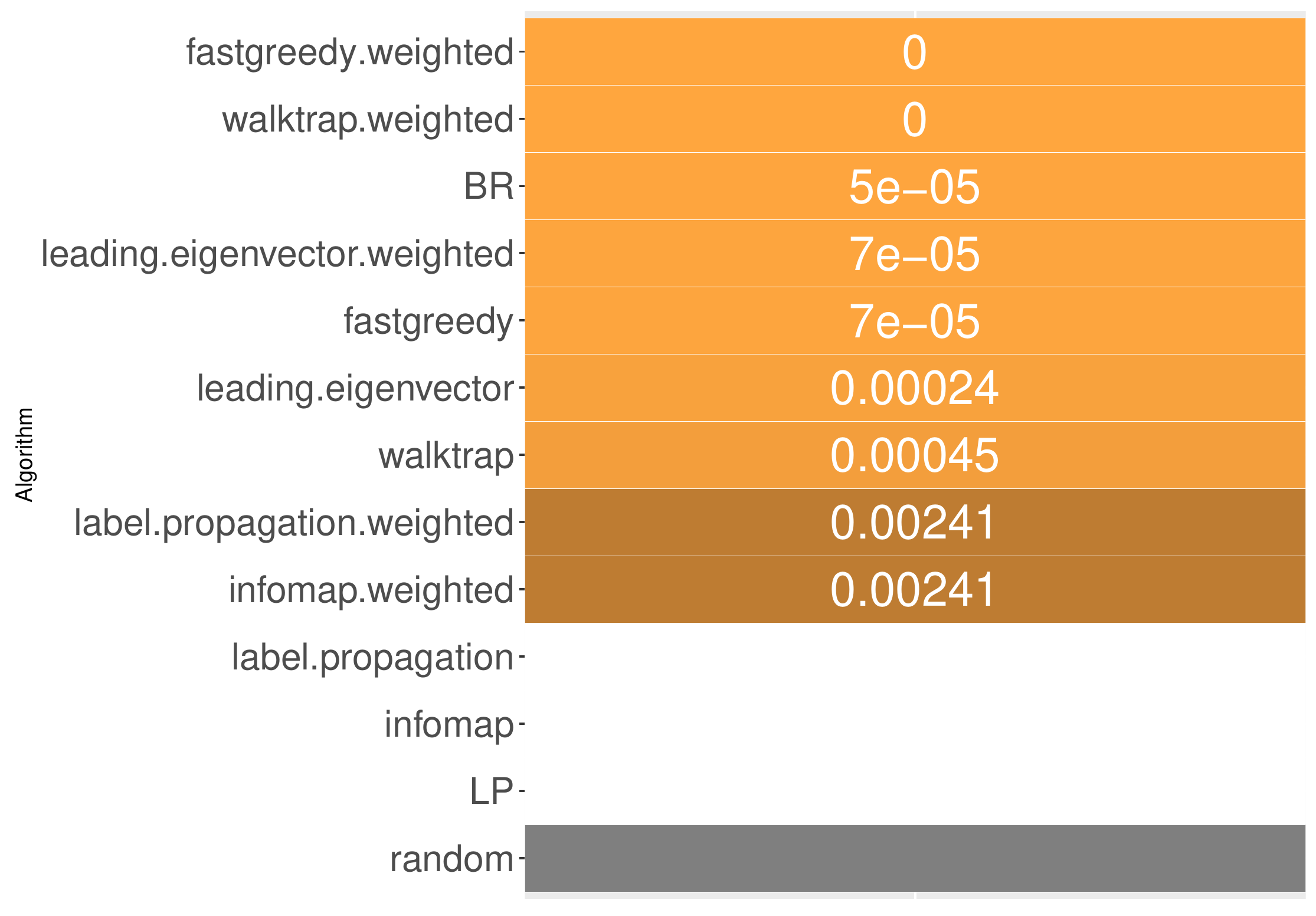}
  \end{center}
\caption{Statistical evaluation of method's performance in terms of micro averaged F1 score. Gray - baseline, white - statistically identical to baseline, otherwise, the p-value of hypothesis that a method performs better than the baseline.}
\label{fig:f1micro-statistics}
\end{wrapfigure}

When it comes to ranking of how well methods performed in micro averaged F1, fast gredy and walktrap approaches used on weighted label co-occurence graph performed best, followed by BR, leading eigenvector and unweighted walktrap / modularity-maximizations methods. Also weighted label propagation and infomap were statistically significantly better than the average random performance. We confirm \textbf{RH1}.

In terms of micro-averaged F1 weighted fast greedy approach has both the highest mean (86\%) and median (92\%) likelihood of scoring better than random baseline. Binary Relevance and weighted variants of walktrap, leading eigenvector also performed well with a mean likelihood of 83-85\% and a lower, but still satisfactory median of 85-88\%.  We confirm \textbf{RH2}.

Modularity-based approaches also turn out to be most resilient. The weighted variant of walktrap was the most resilient with a 69.5\% likelihood in the worst case, followed closely by a weighted fast greedy approach with 67\% and unweighted walktrap with 66.7\%. We note that, apart from a single outlying data sets, all methods (apart from Label Powerset) had better than 50\% likelihood of performing better than \rakeldstop Binary Relevance worst case likelihood was exactly $0.5$. We this confirm both \textbf{RH3} and \textbf{RH4}.

Fast greedy and walktrap weighted approaches yielded the best advantage over \rakeld both in average and worst cases. Binary Relevance also provided a strong overhead against random label space division, while achieving just $0.5$ in the worst case scenario. Thus, when it comes to micro-averaged F1 scores \rakeld random approaches to label space partitions should be dropped in favor of weighted fast greedy and walktrap methods or Binary Relevance. All of these methods are also statistically significantly better than the average random baseline. We therefore confirm \textbf{RH1}, \textbf{RH2}, \textbf{RH3}, \textbf{RH4} for micro-averaged F1 scores.

\begin{figure}[H]
\centering
\includegraphics[width=.9\textwidth]{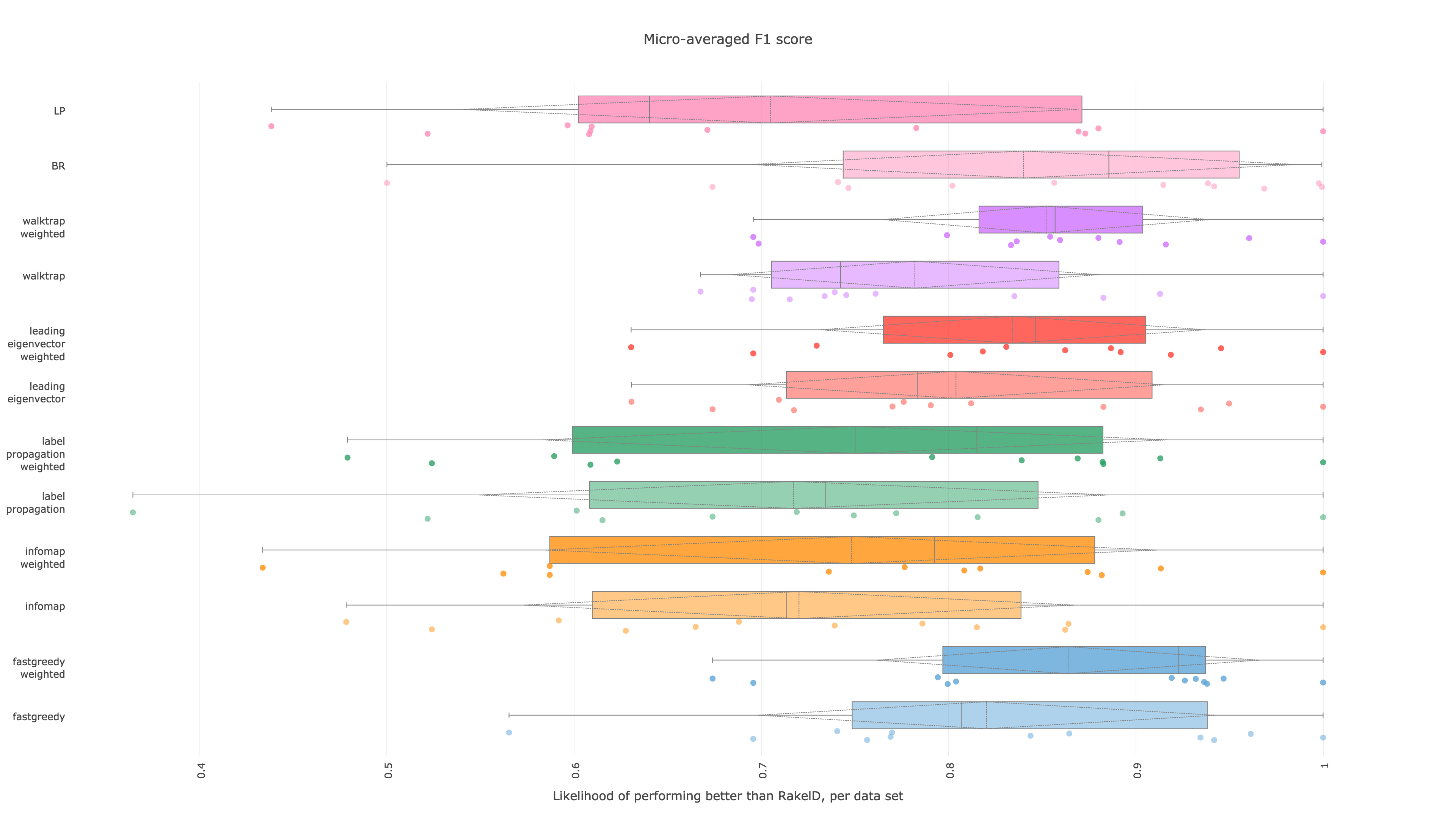}
\caption{Histogram of methods' likelihood  of  performing  better  than  RA$k$EL$d$  in  Micro-averaged  F1  score aggregated over data sets}
\label{fig:f1microhist}
\end{figure}

We also note that \rakeld was better than Label Powerset on micro-averaged F1 in 57\% of cases in the worst case while Tsoumakas et. al.'s original paper \cite{tsoumakas_random_2011} provides argumentation of micro-F1 improvements over LP yielded by \rakeld, using SVMs. We note that our observation is not contrary - LP failed to produce significantly different results than the averange random baseline in our setting. Instead our results are complementary, as we use a different classifier but the intuition can be used to comment on Tsoumakas et. al.'s results. While in some cases \rakeld provides an improvement over LP in F1 score, on average the probability of drawing a random subspace better is only 30\%. We still note that it is much better to use one of the recommended community detection based approaches instead of a method based on a priori assumptions.

\begin{table}[H]
    \centering
    \begin{adjustbox}{width=1\textwidth}
    \small

\begin{tabular}{lrrrrrrrrrrrr}
\toprule
{} &        BR &        LP &  fastgreedy &  fastgreedy-weighted &   infomap &  infomap-weighted &  label\_propagation &  label\_propagation-weighted &  leading\_eigenvector &  leading\_eigenvector-weighted &  walktrap &  walktrap-weighted \\
\midrule
Corel5k     &  0.856444 &  0.608000 &    0.961333 &             0.804000 &  0.524000 &          0.881778 &           0.601333 &                    0.524000 &             0.949778 &                      0.818222 &  0.745333 &           0.799111 \\
bibtex      &  0.997778 &  0.782667 &    0.756444 &             0.794222 &  0.664889 &          0.816889 &           0.749333 &                    0.882222 &             0.812000 &                      0.800889 &  0.835111 &           0.833333 \\
birds       &  0.968562 &  0.438280 &    0.843736 &             0.946833 &  0.591771 &          0.433657 &           0.364309 &                    0.478964 &             0.630606 &                      0.830791 &  0.694868 &           0.836338 \\
delicious   &  0.914667 &  0.869333 &    0.941778 &             0.936444 &  0.864000 &          0.874222 &           0.892889 &                    0.868889 &             0.934667 &                      0.918667 &  0.912889 &           0.916000 \\
emotions    &  0.500000 &  0.521739 &    0.565217 &             0.673913 &  0.739130 &          0.586957 &           0.673913 &                    0.913043 &             0.717391 &                      0.630435 &  0.739130 &           0.891304 \\
enron       &  0.802000 &  0.873000 &    0.934500 &             0.938000 &  0.786000 &          0.776500 &           0.815500 &                    0.839000 &             0.776000 &                      0.945500 &  0.761000 &           0.859500 \\
genbase     &  0.941778 &  0.880000 &    0.864444 &             0.919111 &  0.862222 &          0.913333 &           0.880000 &                    0.882667 &             0.882667 &                      0.862222 &  0.882667 &           0.880000 \\
mediamill   &  0.740889 &  0.609333 &    0.769778 &             0.932000 &  0.627556 &          0.562222 &           0.615111 &                    0.589333 &             0.709333 &                      0.886667 &  0.715111 &           0.854222 \\
medical     &  0.938500 &  0.596500 &    0.769000 &             0.799500 &  0.688000 &          0.736000 &           0.772000 &                    0.623000 &             0.770000 &                      0.729500 &  0.667500 &           0.698500 \\
scene       &  0.673913 &  0.608696 &    0.695652 &             0.695652 &  0.478261 &          0.586957 &           0.521739 &                    0.608696 &             0.673913 &                      0.695652 &  0.695652 &           0.695652 \\
tmc2007-500 &  0.999343 &  1.000000 &    1.000000 &             1.000000 &  1.000000 &          1.000000 &           1.000000 &                    1.000000 &             1.000000 &                      1.000000 &  1.000000 &           1.000000 \\
yeast       &  0.746458 &  0.671141 &    0.740492 &             0.926174 &  0.815063 &          0.808352 &           0.718867 &                    0.791201 &             0.790455 &                      0.891872 &  0.733781 &           0.960477 \\
\bottomrule
\end{tabular}

\end{adjustbox}
\label{tab:f1-micro.all}
\caption{Likelihood of performing better than RA$k$EL$d$ in Micro-averaged F1 score of every method for each data set}
\end{table} 
\subsection{Macro-averaged F1 score}

\begin{wrapfigure}{L}{0.33\textwidth}
  \begin{center}
    \includegraphics[width=.3\textwidth]{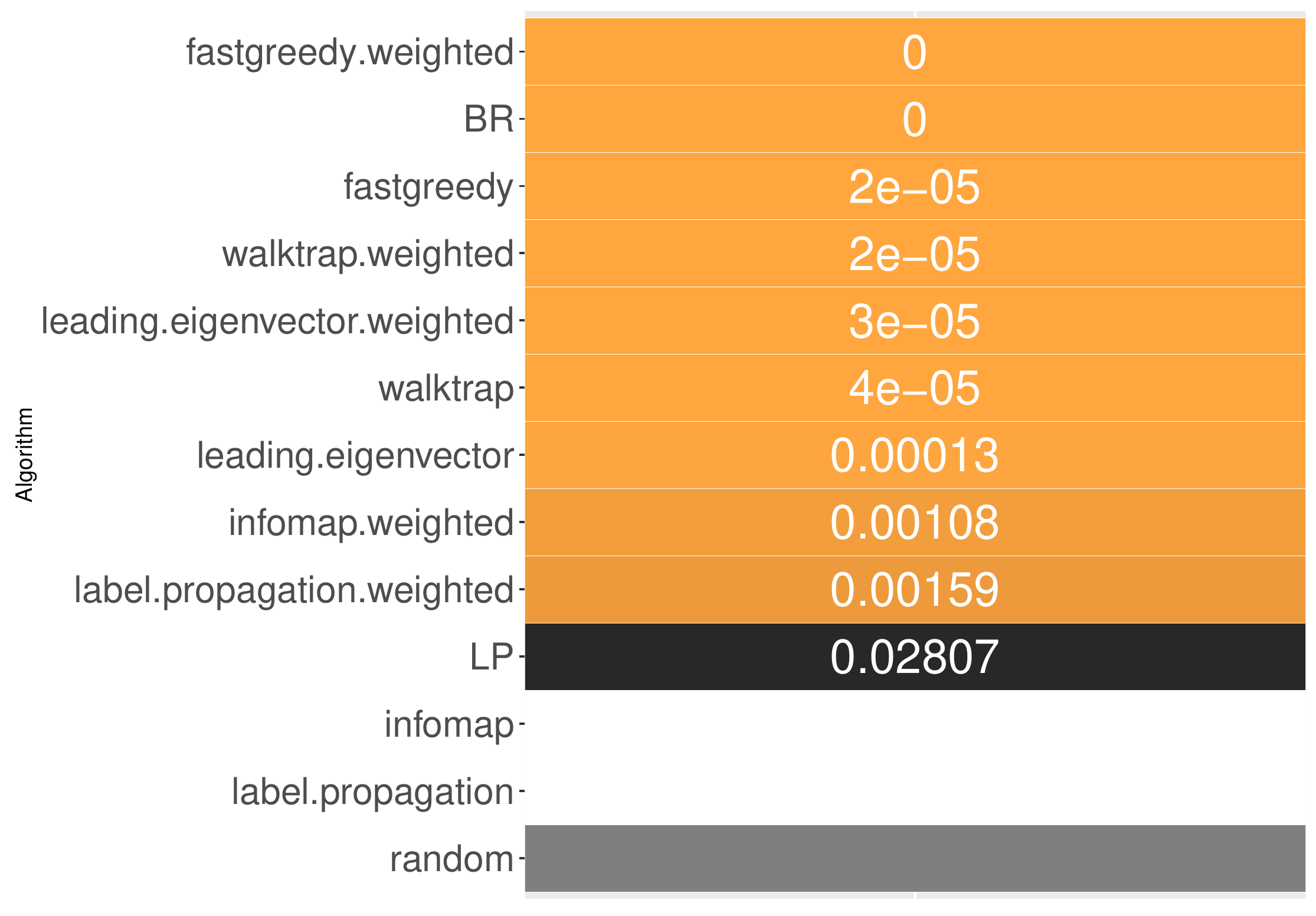}
  \end{center}
\caption{Statistical evaluation of method's performance in terms of macro averaged F1 score. Gray - baseline, white - statistically identical to baseline, otherwise, the p-value of hypothesis that a method performs better than the baseline.}
\label{fig:f1macro-statistics}
\end{wrapfigure}

All methods, apart from unweighted label propagation and infomap, performed significantly better than the average random baseline. The highest ranks were achieved by weighted fast greedy, Binary Relevance and unweighted fast greedy. We confirm \textbf{RH1}.

Fast gredy and walktrap approaches used on weighted label co-occurence graph were most likely to perform better than \rakeld samples, followed by BR, leading eigenvector and unweighted walktrap / modularity-maximizations methods. Also weighted label propagation and infomap were statistically significantly better than the average random performance.

Binary Relevance, weighted fast greedy were the two approaches that surpassed the 90\% likelihood of being better than random label space divisions in both median (98.5\% and 97\% respectively) and mean (92\% and 90\%) cases. Weighted walktrap and leading eigenvector followed closely with both the median and the mean likelihood of 87-89\%. We confirm thus reject \textbf{RH2} as Binary Relevance achieved greater likelihoods than the best data-driven approach.

When it comes to resilience - all modularity (apart from unweighted fast greedy) methods, achieve the same high worst-case 70\% probability of performing better than \rakeldstop Binary Relevance underperformed in worst-case with being better exactly in 50\% of the cases. All methods on all data sets, apart from outlier case of infomap's and label propagation's performance on the scene data set, are likely to yield a better macro-averaged F1 score than the random approaches. We confirm \textbf{RH3} and \textbf{RH4}.

We recommend using Binary Relevance or weighted fast greedy approaches when generalizing to achieve best macro-averaged F1 score, as they are both significantly better than average random performance, more likely to perform better than \rakeld samplings and this likelihood is high even in the worst case. Thus for macro-averaged F1 we confirm hypotheses \textbf{RH1}, \textbf{RH3}, \textbf{RH4}. Binary Relevance had a slightly better likelihood of beating \rakeld than data-driven approaches and thus we reject \textbf{RH2}.

\begin{figure}[H]
\centering
\includegraphics[width=.9\textwidth]{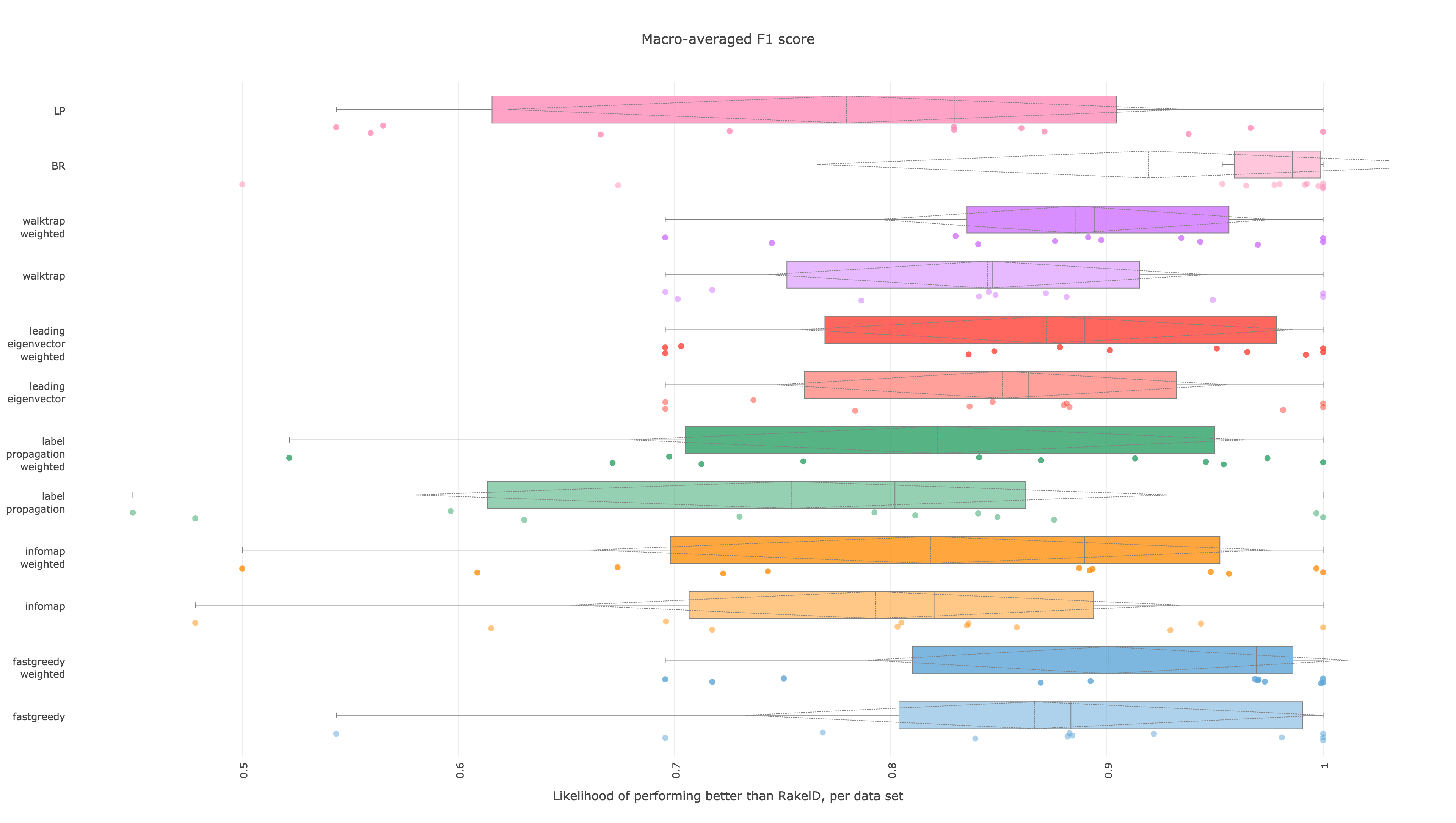}
\caption{Histogram of methods' likelihood  of  performing  better  than  RA$k$EL$d$  in  Macro-averaged  F1  score aggregated over data sets}
\label{fig:f1macrohist}
\end{figure}

\begin{table}[h]
    \centering
    \begin{adjustbox}{width=1\textwidth}
    \small

\begin{tabular}{lrrrrrrrrrrrr}
\toprule
{} &        BR &        LP &  fastgreedy &  fastgreedy-weighted &   infomap &  infomap-weighted &  label\_propagation &  label\_propagation-weighted &  leading\_eigenvector &  leading\_eigenvector-weighted &  walktrap &  walktrap-weighted \\
\midrule
Corel5k     &  1.000000 &  0.665778 &    0.980889 &             0.968444 &  0.615111 &          0.996889 &           0.596444 &                    0.712444 &             0.836444 &                      0.901333 &  0.845333 &           0.969778 \\
bibtex      &  1.000000 &  0.871111 &    0.839111 &             0.869333 &  0.803111 &          0.887111 &           0.849333 &                    0.945778 &             0.880000 &                      0.878222 &  0.881333 &           0.876000 \\
birds       &  0.992603 &  0.559408 &    0.883957 &             0.999075 &  0.804901 &          0.673601 &           0.449376 &                    0.671290 &             0.736477 &                      0.847896 &  0.786408 &           0.897365 \\
delicious   &  0.997778 &  0.937778 &    1.000000 &             1.000000 &  0.929333 &          0.956444 &           0.996889 &                    0.974222 &             1.000000 &                      1.000000 &  1.000000 &           1.000000 \\
emotions    &  0.500000 &  0.543478 &    0.543478 &             0.717391 &  0.717391 &          0.608696 &           0.630435 &                    0.913043 &             0.695652 &                      0.695652 &  0.717391 &           0.891304 \\
enron       &  0.991500 &  0.966500 &    1.000000 &             0.973000 &  0.943500 &          0.948000 &           0.875500 &                    0.954000 &             0.981500 &                      0.992000 &  0.949000 &           0.830000 \\
genbase     &  0.953333 &  0.829333 &    0.881778 &             0.892444 &  0.836000 &          0.892000 &           0.840444 &                    0.840889 &             0.882667 &                      0.836000 &  0.848444 &           0.840444 \\
mediamill   &  0.964444 &  0.860444 &    0.882667 &             0.969778 &  0.835111 &          0.743111 &           0.792444 &                    0.759556 &             0.881333 &                      0.964889 &  0.840889 &           0.943111 \\
medical     &  0.977500 &  0.725500 &    0.768500 &             0.750500 &  0.696000 &          0.722500 &           0.730000 &                    0.697500 &             0.783500 &                      0.703000 &  0.701500 &           0.745000 \\
scene       &  0.673913 &  0.565217 &    0.695652 &             0.695652 &  0.478261 &          0.500000 &           0.478261 &                    0.521739 &             0.695652 &                      0.695652 &  0.695652 &           0.695652 \\
tmc2007-500 &  1.000000 &  1.000000 &    1.000000 &             1.000000 &  1.000000 &          1.000000 &           1.000000 &                    1.000000 &             1.000000 &                      1.000000 &  1.000000 &           1.000000 \\
yeast       &  0.979866 &  0.829232 &    0.921700 &             0.970172 &  0.858315 &          0.893363 &           0.811335 &                    0.869500 &             0.847129 &                      0.950783 &  0.871738 &           0.934377 \\
\bottomrule
\end{tabular}

\end{adjustbox}
\label{tab:f1-macro.all}
\caption{Likelihood of performing better than RA$k$EL$d$ in Macro-averaged F1 score of every method for each data set}
\end{table} 

\newpage
\subsection{Subset Accuracy}

\begin{wrapfigure}{l}{0.33\textwidth}
  \begin{center}
    \includegraphics[width=.3\textwidth]{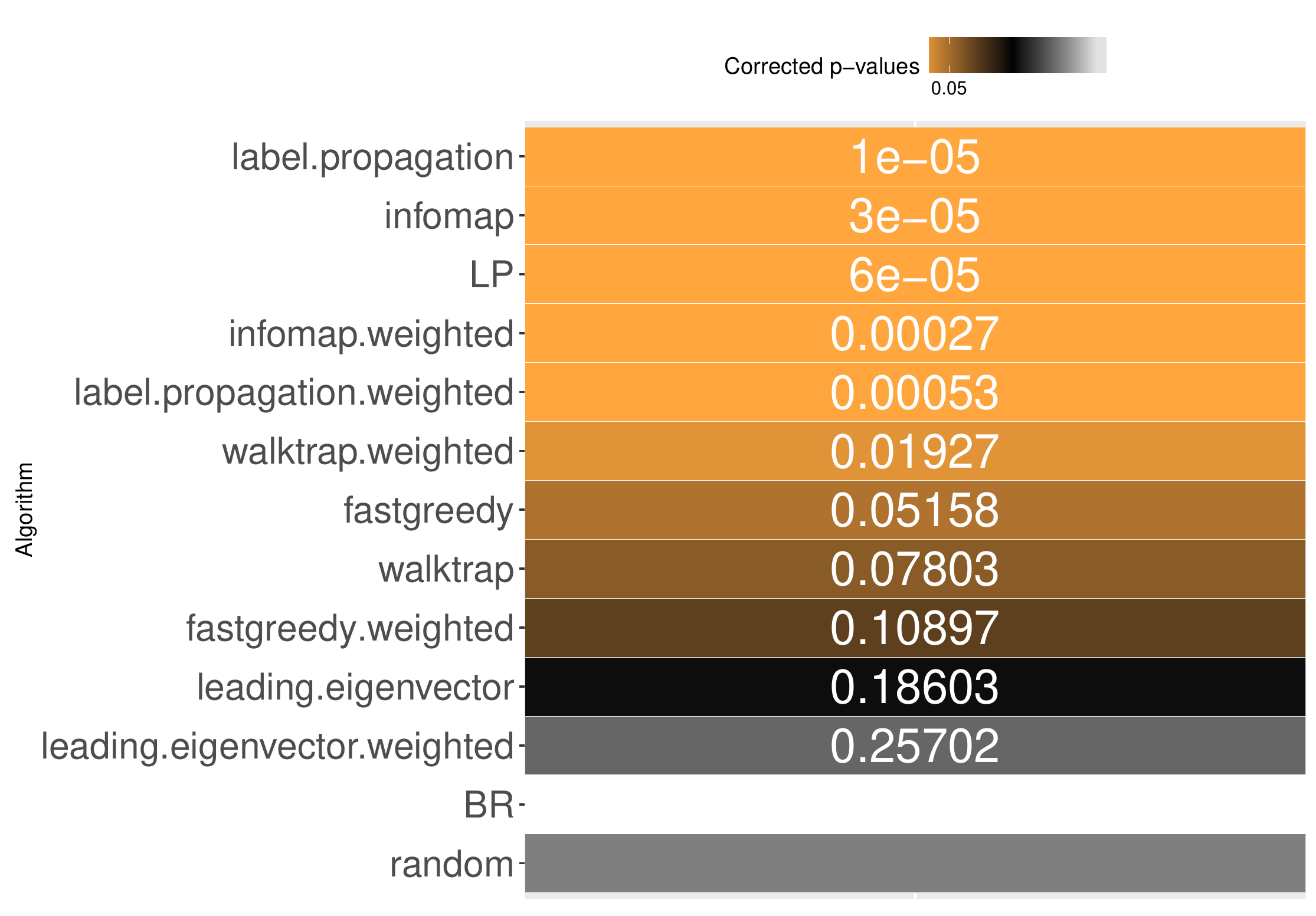}
  \end{center}
\caption{Statistical evaluation of method's performance in terms of Jaccard Similarity score. Gray - baseline, white - statistically identical to baseline, otherwise, the p-value of hypothesis that a method performs better than the baseline.}
\label{fig:sa-statistics}
\end{wrapfigure}

All methods apart from weighted leading eigenvector modularity maximization approach were statistically significantly better than the average random baseline. 

Label propagation, infomap, Label Powerset, weighted infomap, label propagation and walktrap are the methods that performed statistically significantly better than average random baseline, ordered by ranks. We confirm \textbf{RH1}.

Also unweighted infomap and label propagation are the most likely to yield results of higher subset accuracy than random label space divisions, both regarding the median (96\%) and the mean (90-91\%) likelihood. Label Powerset follows with a 95.8\% median and 89\% mean. Weighted versions of infomap and label powerset are fourth and fifth with 5-6 percentage points less. We confirm \textbf{RH2}.

Concerning resiliency of the advantage, only infomap versions proved to be better than \rakeld in more than half of the times - the unweighted version in 58\% of cases, the weighted one in 52\%. All other methods were below the 50\% threshold in the worst case, with Label Powerset and label propagation likelihood of 33\% for both variants. If one or two most wrong outliers were to be discarded, all methods are more than 50\% likely to be better than random label space partitioning. We confirm \textbf{RH3} and \textbf{RH4}.

\begin{figure}[H]
\centering
\includegraphics[width=.9\textwidth]{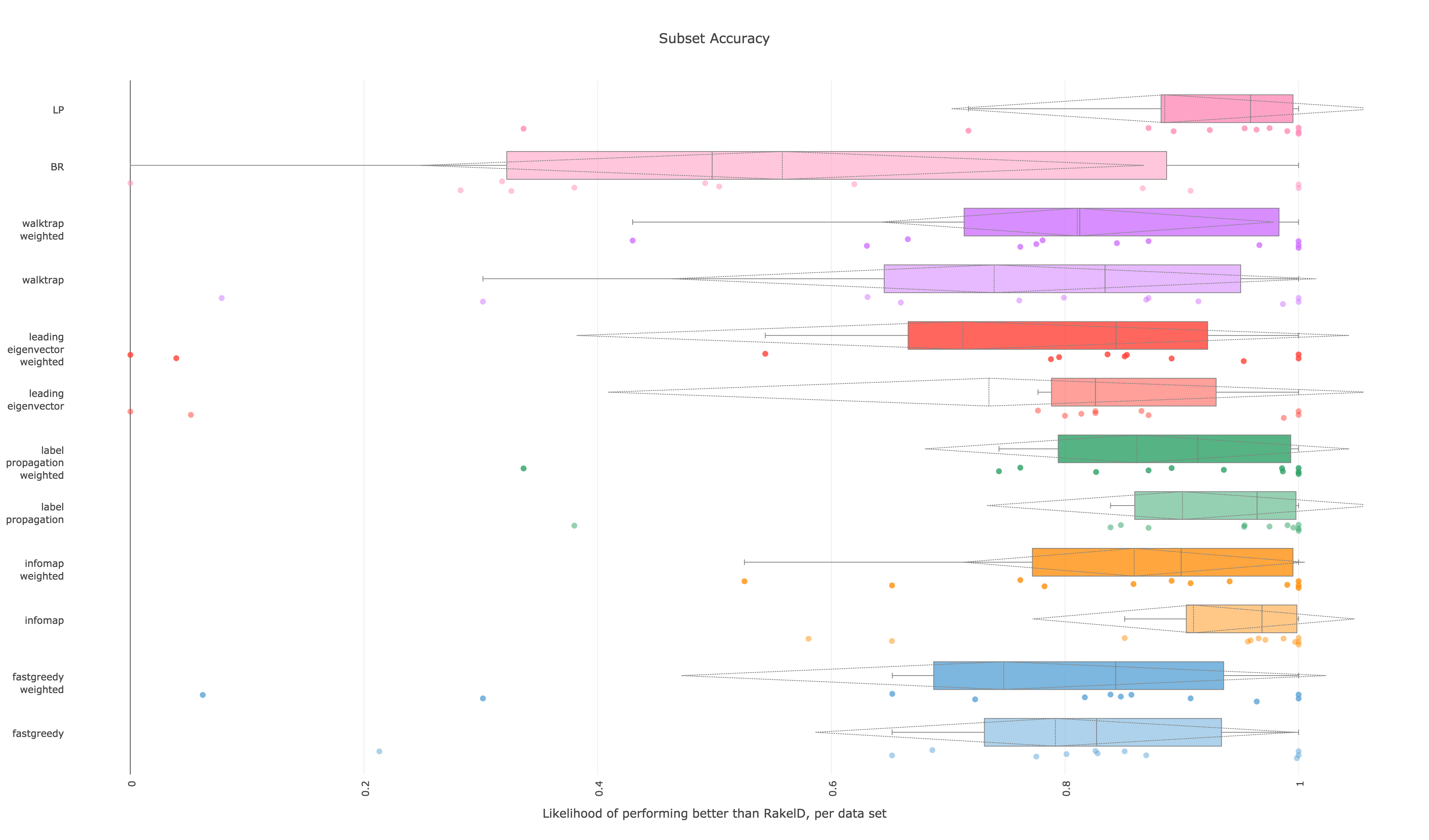}
\caption{Histogram of methods' likelihood  of  performing  better  than  RA$k$EL$d$  in  Subset Accuracy aggregated over data sets}
\label{fig:sahist}
\end{figure}

We thus recommend using unweighted infomap as the data-driven alternative to \rakeld, as it is both significantly better than the random baseline, very likely to perform batter than \rakeld and most resilient among evaluated methods in the worst case. We confirm \textbf{RH1}, \textbf{RH2}, \textbf{RH3} and \textbf{RH4} for subset accuracy.

\begin{table}[h]
    \centering
    \begin{adjustbox}{width=1\textwidth}
    \small

\begin{tabular}{lrrrrrrrrrrrr}
\toprule
{} &        BR &        LP &  fastgreedy &  fastgreedy-weighted &   infomap &  infomap-weighted &  label\_propagation &  label\_propagation-weighted &  leading\_eigenvector &  leading\_eigenvector-weighted &  walktrap &  walktrap-weighted \\
\midrule
Corel5k     &  0.000000 &  0.953778 &    0.652000 &             0.301778 &  0.965778 &          0.652000 &           0.953778 &                    0.826667 &             0.000000 &                      0.000000 &  0.301778 &           0.780889 \\
bibtex      &  0.492000 &  0.975111 &    0.828000 &             0.723111 &  0.971556 &          0.761778 &           0.975111 &                    0.761778 &             0.800000 &                      0.788000 &  0.799111 &           0.761778 \\
birds       &  0.380028 &  0.336570 &    0.213130 &             0.061951 &  0.651872 &          0.525659 &           0.380028 &                    0.336570 &             0.051780 &                      0.039297 &  0.078132 &           0.429958 \\
delicious   &  1.000000 &  1.000000 &    1.000000 &             1.000000 &  1.000000 &          1.000000 &           1.000000 &                    1.000000 &             1.000000 &                      1.000000 &  1.000000 &           1.000000 \\
emotions    &  0.326087 &  0.717391 &    0.826087 &             0.847826 &  0.956522 &          0.891304 &           0.847826 &                    0.891304 &             0.826087 &                      0.891304 &  0.760870 &           1.000000 \\
enron       &  0.504000 &  0.964000 &    0.775500 &             0.817000 &  0.959000 &          0.941000 &           0.990500 &                    0.986500 &             0.865500 &                      0.795000 &  0.659500 &           0.775500 \\
genbase     &  0.907556 &  0.871556 &    0.851111 &             0.907556 &  0.851111 &          0.907556 &           0.871556 &                    0.871556 &             0.871556 &                      0.851111 &  0.871556 &           0.871556 \\
mediamill   &  0.318222 &  0.924000 &    0.801333 &             0.856889 &  0.987111 &          0.858667 &           0.953333 &                    0.936000 &             0.776889 &                      0.836444 &  0.914222 &           0.844444 \\
medical     &  0.866500 &  0.893000 &    0.686500 &             0.839000 &  0.580500 &          0.782500 &           0.839000 &                    0.743500 &             0.814000 &                      0.853000 &  0.631000 &           0.665500 \\
scene       &  0.282609 &  1.000000 &    0.869565 &             0.652174 &  1.000000 &          1.000000 &           1.000000 &                    1.000000 &             0.826087 &                      0.543478 &  0.869565 &           0.630435 \\
tmc2007-500 &  1.000000 &  1.000000 &    1.000000 &             1.000000 &  1.000000 &          1.000000 &           1.000000 &                    1.000000 &             1.000000 &                      1.000000 &  1.000000 &           1.000000 \\
yeast       &  0.619687 &  0.990306 &    0.998509 &             0.964206 &  0.997017 &          0.990306 &           0.995526 &                    0.985831 &             0.987323 &                      0.953020 &  0.986577 &           0.966443 \\
\bottomrule
\end{tabular}

\end{adjustbox}
\label{tab:accuracy.all}
\caption{Likelihood of performing better than RA$k$EL$d$ in Subset Accuracy of every method for each data set}
\end{table} 

\subsection{Jaccard score}

\begin{figure}[H]
\centering
\includegraphics[width=.9\textwidth]{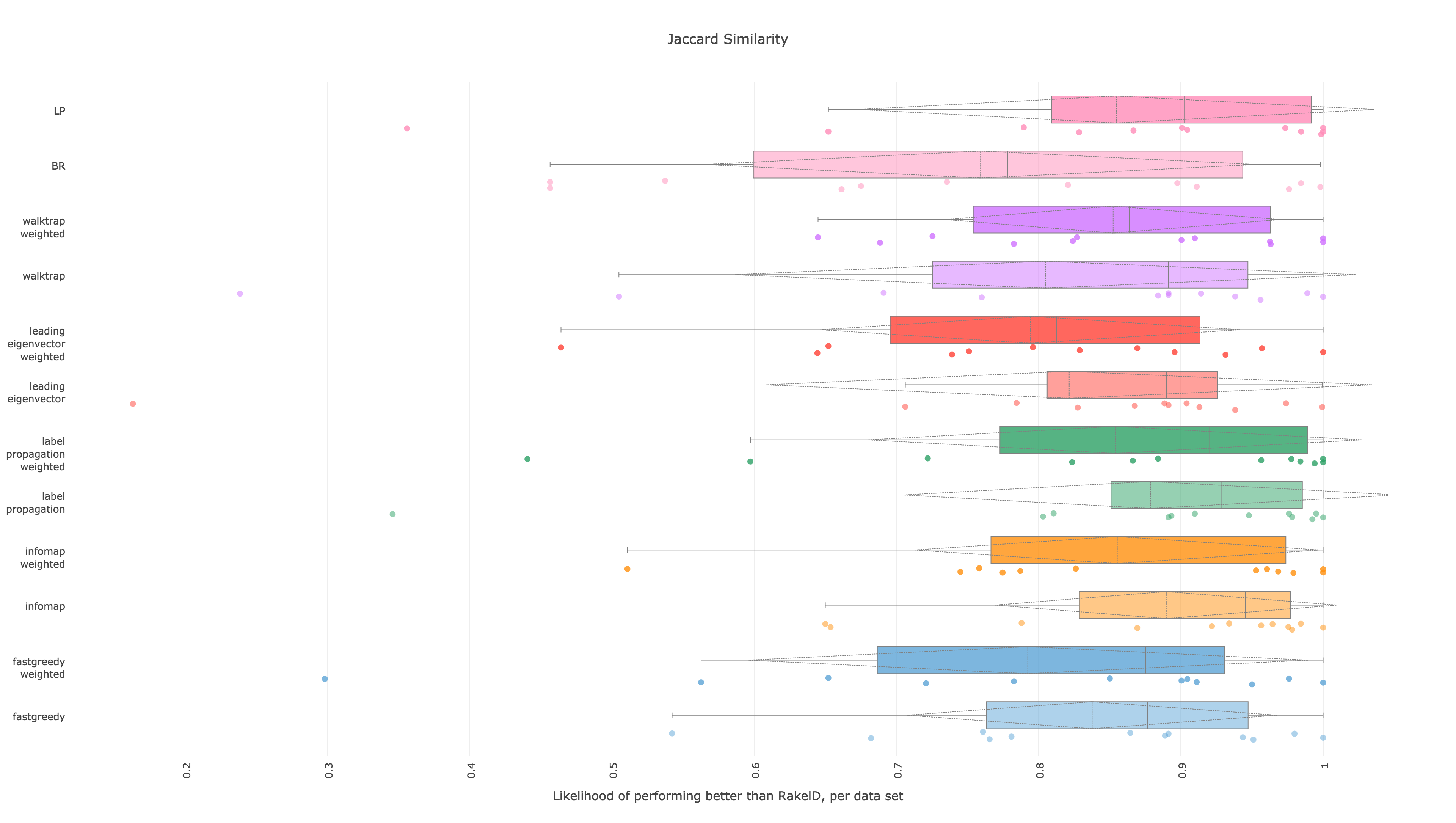}
\caption{Histogram of methods' likelihood  of  performing  better  than  RA$k$EL$d$  in  Jaccard Similarity aggregated over data sets}
\label{fig:jshist}
\end{figure}

\begin{wrapfigure}{L}{0.33\textwidth}
  \begin{center}
    \includegraphics[width=.3\textwidth]{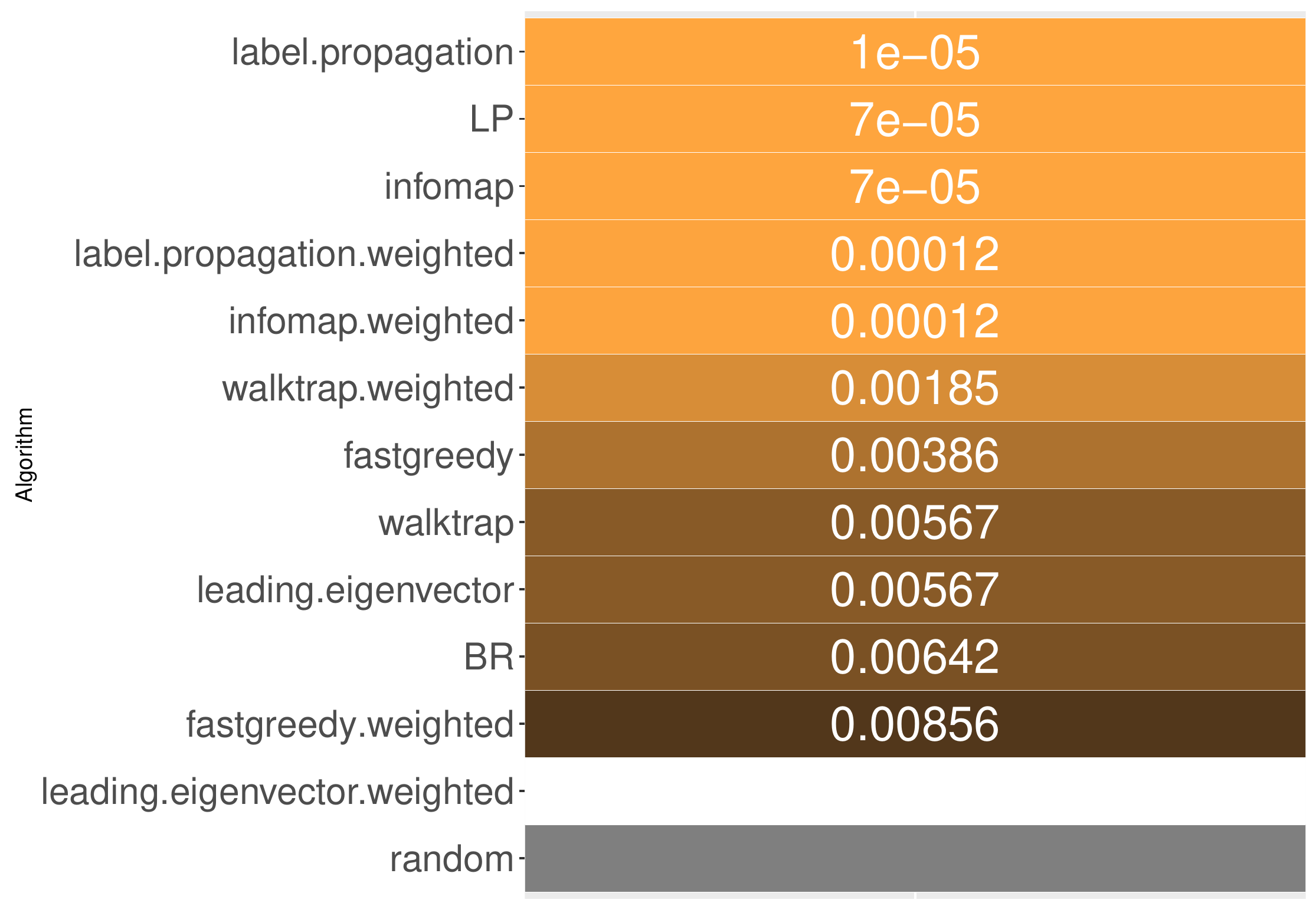}
  \end{center}
\caption{Statistical evaluation of method's performance in terms of micro averaged F1 score. Gray - baseline, white - statistically identical to baseline, otherwise, the p-value of hypothesis that a method performs better than the baseline.}
\label{fig:js-statistics}
\end{wrapfigure}

All methods apart from weighted leading eigenvector modularity maximization approach were statistically significantly better than the average random baseline. Unweighted label propagation, Label Powerset and infomap were the highest ranked method. We confirm \textbf{RH1}.

Jaccard score is similar to Subset Accuracy in rewarding exact label set matches. In effect, it is not surprising to see that unweighted infomap and label propagation are the most likely than \rakeld to yield a result of higher Jaccard both in terms of median (94.5\% and 92.9\% respectively) and mean likelihoods (88.9\% and 87.9\% resp.). Out of the two, infomap provides the most resilient advantage with a 65\% probability of performing better than random approaches in the worst case. Label propagation is in worst cases only 34-35\% likely to be better than random space partitions.

We recommend using unweighted infomap approach over \rakeld when Jaccard similarity is of importance and confirm \textbf{RH1}, \textbf{RH2}, \textbf{RH3} and \textbf{RH4} for this measure.

\begin{table}[ht]
    \centering
    \begin{adjustbox}{width=1\textwidth}
    \small

\begin{tabular}{lrrrrrrrrrrrr}
\toprule
{} &        BR &        LP &  fastgreedy &  fastgreedy-weighted &   infomap &  infomap-weighted &  label\_propagation &  label\_propagation-weighted &  leading\_eigenvector &  leading\_eigenvector-weighted &  walktrap &  walktrap-weighted \\
\midrule
Corel5k     &  0.675111 &  0.828444 &    0.888889 &             0.562667 &  0.788000 &          0.787111 &           0.803111 &                    0.597333 &             0.888444 &                      0.644444 &  0.504889 &           0.644889 \\
bibtex      &  0.984444 &  0.998667 &    0.780889 &             0.720889 &  0.975556 &          0.758222 &           0.995111 &                    0.866222 &             0.867556 &                      0.796000 &  0.938222 &           0.824000 \\
birds       &  0.820620 &  0.355987 &    0.542302 &             0.298197 &  0.653722 &          0.510865 &           0.345816 &                    0.440592 &             0.163199 &                      0.464170 &  0.238558 &           0.725381 \\
delicious   &  0.976000 &  0.973333 &    0.943556 &             0.900444 &  0.964444 &          0.968444 &           0.992444 &                    0.984000 &             0.938222 &                      0.828889 &  0.988889 &           0.963111 \\
emotions    &  0.456522 &  0.652174 &    0.760870 &             0.652174 &  0.956522 &          0.826087 &           0.891304 &                    0.956522 &             0.891304 &                      0.652174 &  0.891304 &           1.000000 \\
enron       &  0.735500 &  0.984500 &    0.951000 &             0.904500 &  0.934000 &          0.960500 &           0.976000 &                    0.994000 &             0.784500 &                      0.957000 &  0.760000 &           0.827000 \\
genbase     &  0.911111 &  0.904444 &    0.864444 &             0.911111 &  0.869333 &          0.952889 &           0.909778 &                    0.884000 &             0.904000 &                      0.869333 &  0.884000 &           0.909778 \\
mediamill   &  0.537333 &  0.866667 &    0.682222 &             0.976000 &  0.921778 &          0.774667 &           0.893333 &                    0.823556 &             0.706222 &                      0.895556 &  0.914222 &           0.900444 \\
medical     &  0.897500 &  0.789500 &    0.765500 &             0.850000 &  0.650000 &          0.745000 &           0.810500 &                    0.722000 &             0.827500 &                      0.751000 &  0.691000 &           0.688500 \\
scene       &  0.456522 &  1.000000 &    0.891304 &             0.782609 &  0.978261 &          1.000000 &           0.978261 &                    1.000000 &             0.913043 &                      0.739130 &  0.891304 &           0.782609 \\
tmc2007-500 &  0.998029 &  1.000000 &    1.000000 &             1.000000 &  1.000000 &          1.000000 &           1.000000 &                    1.000000 &             0.999343 &                      1.000000 &  1.000000 &           1.000000 \\
yeast       &  0.661447 &  0.900820 &    0.979866 &             0.950037 &  0.984340 &          0.979120 &           0.947800 &                    0.977629 &             0.973900 &                      0.931394 &  0.956003 &           0.962714 \\
\bottomrule
\end{tabular}

\end{adjustbox}
\label{tab:jaccard.all}
\caption{Likelihood of performing better than RA$k$EL$d$ in Jaccard Similarity of every method for each data set}
\end{table} 

\subsection{Hamming Loss}

\begin{figure}[h]
\centering
\includegraphics[width=.9\textwidth]{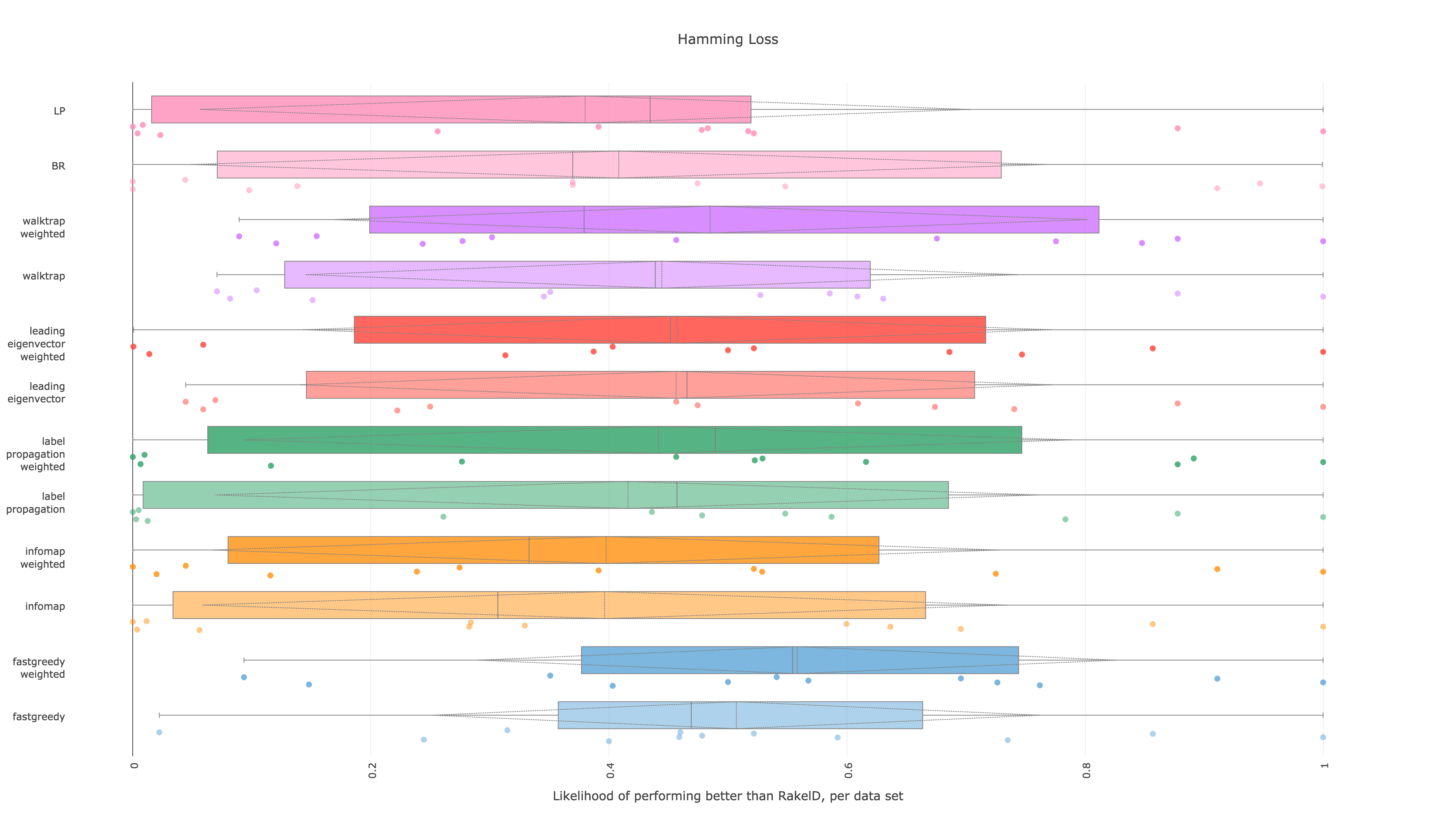}
\caption{Histogram of methods' likelihood  of  performing  better  than  RA$k$EL$d$  in Hamming Loss Similarity aggregated over data sets}
\label{fig:hshist}
\end{figure}

Hamming Loss is certainly a fascinating case in our experiments. As the measure is evaluated per each label separately, we can expect it to be the most stable over different label space partitions. 

The first surprise comes with Friedman-Iman-Davenport test result, where the test practically fails to find a difference in performance between random approaches and data-driven methods, yielding a p-value of $0.049$. While the p-value is lower than $\alpha = 0.05$, but the difference cannot be taken as significant given the characteristics of the test. Lack of significance is confirmed by pairwise tests against random baseline (all hypotheses of difference are strongly rejected). We reject \textbf{RH1}.

Weighted fast greedy was the only approach to be more likely to yield a lower Hamming Loss than \rakeld on average, both in median and mean (55\%) likelihoods. The unweighted version was better than slightly over 50\% more cases than \rakeld, with a median likelihood of 46\%. Binary Relevance and Label Powerset achieved likelihoods lower by close to 10 percentage points. We thus confirm \textbf{RH2}.

When it comes to worst-case observations Binary Relevance, Label Powerset and infomap in both variants were never better than \rakeldstop The methods with the most resilient advantage in likelihood (9\%) in the worst case were weighted versions of fastgreedy and walktrap. We confirm \textbf{RH3} and reject \textbf{RH4}.

We conclude that fast greedy approach can be recommended over \rakeld, as even given such large standard deviation of likelihoods it still yields lower Hamming Loss than random label space divisions on more than half of the data sets. Yet we reject \textbf{RH1},  \textbf{RH2} and \textbf{RH4} for Hamming Loss. We confirm \textbf{RH3} as a priori methods do not provide better performance than \rakeld in the worst case.

\begin{table}[h]
    \centering
    \begin{adjustbox}{width=1\textwidth}
    \small

\begin{tabular}{lrrrrrrrrrrrr}
\toprule
{} &        BR &        LP &  fastgreedy &  fastgreedy-weighted &   infomap &  infomap-weighted &  label\_propagation &  label\_propagation-weighted &  leading\_eigenvector &  leading\_eigenvector-weighted &  walktrap &  walktrap-weighted \\
\midrule
Corel5k     &  0.000000 &  0.004000 &    0.400000 &             0.148000 &  0.003556 &          0.115556 &           0.004889 &                    0.009778 &             0.069333 &                      0.000444 &  0.350667 &           0.243556 \\
bibtex      &  0.097778 &  0.023111 &    0.022222 &             0.093333 &  0.011556 &          0.044444 &           0.012444 &                    0.116000 &             0.059111 &                      0.059111 &  0.070667 &           0.089333 \\
birds       &  0.474341 &  0.008322 &    0.459085 &             0.540915 &  0.055941 &          0.019880 &           0.002774 &                    0.006472 &             0.044383 &                      0.013870 &  0.104022 &           0.154415 \\
delicious   &  0.000000 &  0.000000 &    0.244444 &             0.350667 &  0.000000 &          0.000000 &           0.000000 &                    0.000000 &             0.249778 &                      0.387111 &  0.081778 &           0.120444 \\
emotions    &  0.369565 &  0.391304 &    0.478261 &             0.695652 &  0.695652 &          0.521739 &           0.586957 &                    0.891304 &             0.673913 &                      0.521739 &  0.608696 &           0.847826 \\
enron       &  0.044000 &  0.483000 &    0.460000 &             0.567500 &  0.284000 &          0.274500 &           0.436000 &                    0.522500 &             0.474500 &                      0.313000 &  0.151000 &           0.277000 \\
genbase     &  0.911111 &  0.877778 &    0.856889 &             0.911111 &  0.856889 &          0.911111 &           0.877778 &                    0.877778 &             0.877778 &                      0.856889 &  0.877778 &           0.877778 \\
mediamill   &  0.138222 &  0.256000 &    0.314667 &             0.403111 &  0.329333 &          0.238667 &           0.260889 &                    0.276444 &             0.222222 &                      0.403111 &  0.345333 &           0.301778 \\
medical     &  0.947000 &  0.517000 &    0.735000 &             0.762000 &  0.636500 &          0.725000 &           0.783500 &                    0.529000 &             0.740500 &                      0.747000 &  0.585500 &           0.675500 \\
scene       &  0.369565 &  0.521739 &    0.521739 &             0.500000 &  0.282609 &          0.391304 &           0.478261 &                    0.456522 &             0.456522 &                      0.500000 &  0.630435 &           0.456522 \\
tmc2007-500 &  0.999343 &  1.000000 &    1.000000 &             1.000000 &  1.000000 &          1.000000 &           1.000000 &                    1.000000 &             1.000000 &                      1.000000 &  1.000000 &           1.000000 \\
yeast       &  0.548098 &  0.478001 &    0.592095 &             0.726324 &  0.599553 &          0.528710 &           0.548098 &                    0.615958 &             0.609247 &                      0.686055 &  0.527218 &           0.775541 \\
\bottomrule
\end{tabular}

\end{adjustbox}
\label{tab:hl.all}
\caption{Likelihood of performing better than RA$k$EL$d$ in Hamming Loss of every method for each data set}
\end{table}

\section{Conclusions}
\label{sec:finish}
 
We have compared seven approaches as an alternative to random label space partition. RA$k$EL$d$ served as the random baseline for which we have drawn at most 250 distinct label space partitions for at most ten different values of the parameter $k$ of label subset sizes. Out of the seven methods, five inferred the label space partitioning from training data in the data sets, while the two others were based on a priori assumption on how to divide the label space. We evaluated these methods on 12 well-established benchmark data sets. 

We conclude that in four of five measures: micro-/macro-averaged F1 score, Subset Accuracy, and Jaccard similarity all of our proposed methods were more likely to yield better scores than \rakeld apart from single outlying data sets, a data-driven approach was better than average random baseline with statistical significance at $\alpha = .05$. Data-driven approach was also better than \rakeld in worst case scenarios. Thus hypotheses \textbf{RH1}, \textbf{RH3} and \textbf{RH4} has been successfully confirmed with these measures

When it comes to research hypotheses number two (\textbf{RH2}), we've confirmed that with micro-averaged F1, subset accuracy, hamming loss, and jaccard similarity, the data-driven approaches have a higher likelihood of outperforming \rakeld than a priori methods do. The only exception to this is the case of macro averaged F1 where Binary Relevance was most likely to beat random approaches, while followed closely by a data-driven approach - weighted fast greedy.

Hamming Loss forms a separate case for discussion, as this measures is most unrelated to label groups - it is calculated per label only. With this measure most data-driven methods performed much worse than in other methods. Our study failed to observer a statistical difference between data-driven methods and the random baseline, thus we reject hypothesis \textbf{RH1}. For best performing data-driven methods worst-case likelihood of yielding a lower hamming loss than \rakeld was close to 10\% which is far from a resilient score, thus we also reject \textbf{RH4}. We confirm \textbf{RH2} and \textbf{RH3} as there existed a data-driven approach that performed better than a priori approaches.

All in all, statistical significance of a data-driven approach performing better than the averaged random baseline (\textbf{RH1}) has been confirmed for all measures except Hamming Loss. We have confirmed that data-driven approach was more likely than Binary Relevance/Label Powerset to perform better than \rakeld (\textbf{RH2}) in all measures apart from macro-averaged F1 score, where it followed the best Binary Relevance close. Data-driven were always more likely to outperform \rakeld in the worst case than Binary Relevance/Label Powerset, confirming \textbf{RH3} for all measures. Finally for all measures apart from Hamming Loss data-driven approaches were more likely to outperform \rakeld in the worst case, than otherwise. \textbf{RH4} is thus confirmed for all measures except Hamming Loss.

In case of measures that are label-decomposable the fast greedy community detection approach computed on a weighted label co-occurence graph yielded best results among data driven perspectives and is the recommended choice for F1 measures and Hamming Loss. When the measure is instance-decomposable and not label-decomposable, such as Subset Accuracy or Jaccard Similarity, the infomap algorithm should be used on an unweighted label co-occurence graph.

We conclude that community detection methods offer a viable alternative to both random and a priori assumption-based label space partitioning approches. We summarize our findings in the Table \ref{tab:finalrh}, answering the question in the title of how data driven approach to label space partitioning is likely to perform better than random choice.

\begin{table}[H]
    \begin{adjustbox}{width=1\textwidth}
    \begin{tabular}{p{8cm} p{3cm} p{3cm} p{3cm} p{3cm} p{3cm}}
    \toprule
    Measure                                                                                          & Micro-averaged F1                          & Micro-averaged F1     & Subset Accuracy    & Jaccard Similarity & Hamming Loss          \\ 
    \midrule
    \textbf{RH1}: Data-driven approach is significantly better than random (alpha = 0.05)                          & Yes                                        & Yes                   & Yes                & Yes                & No                    \\
    \textbf{RH2}: Data-driven approach is more likely to outperform RAkELd than a priori methods                   & Yes                                        & No                    & Yes                & Yes                & Yes                    \\
    \textbf{RH3}: Data-driven approach is more likely to outperform RAkELd than a priori methods in the worst case & Yes                                        & Yes                   & Yes                & Yes                & Yes                   \\
    \textbf{RH4}: Data-driven approach is more likely to perform better than RAkELd in the worst case, than otherwise               & Yes                                        & Yes                   & Yes                & Yes                & No                    \\
    \midrule
    Recommended data-driven approach                                                                 & Weighted fast greedy and weighted walktrap & Weighted fast greedy  & Unweighted infomap & Unweighted infomap & Weighted fast greedy  \\
\bottomrule
\end{tabular}
\end{adjustbox}
\caption{The summary of evaluated hypotheses and proposed recommendations of this paper\label{tab:finalrh}}
\end{table}

\acknowledgments{\textbf{Acknowledgments:} The work was partially supported by Fellowship co-financed by European Union within European Social Fund; The European Commission under the 7th Framework Programme, Coordination and Support Action, Grant Agreement Number 316097 [ENGINE]; The National Science Centre the research project 2014-2017 decision no. DEC-2013/09/B/ST6/02317. The authors wish to thank Łukasz Augustyniak for helping with proofreading the article. }


\authorcontributions{\textbf{Author Contributions:} 
Piotr Szymański wrote the experimental code and wrote most of the paper. Tomasz Kajdanowicz helped design experiments, supported interpreting the results, supervised the work, wrote parts of the paper and corrected and proofread the paper.}


\conflictofinterests{\textbf{Conflicts of Interest:} The authors declare no conflict of interest.}


%

\appendix 

\section*{\noindent Appendix: Result Tables}\vspace{6pt} 
\begin{table}[ht]
    \centering
    \small
\begin{tabular}{llrrrrrrrrr}
\toprule
Set Name      &  $k$  &   Number of samplings  \\
\midrule
birds & 17 &  163 \\
emotions & 2  &   15 \\
      & 3  &   10  \\
      & 4  &   15  \\
      & 5  &    6 \\
scene & 2  &   15  \\
      & 3  &   10  \\
      & 4  &   15  \\
      & 5  &    6  \\
tmc2007-500 & 21 &   22  \\
yeast & 12 &   91  \\
\bottomrule
\end{tabular}

    \caption{Number of random samplings from the universum of RakelD label space partitions for cases different then 250 samples.}
    \label{tab:samplings}
    \end{table} 
\begin{table}[H]
    \centering
    \begin{adjustbox}{width=1\textwidth}
    \small

\begin{tabular}{lrrrr}
\toprule
{} &   Minimum &    Median &      Mean &       Std \\
\midrule
BR                           &  0.500000 &  0.885556 &  0.840028 &  0.152530 \\
LP                           &  0.438280 &  0.640237 &  0.704891 &  0.171726 \\
fastgreedy                   &  0.565217 &  0.806757 &  0.820198 &  0.127463 \\
fastgreedy-weighted          &  0.673913 &  \textbf{0.922643} &  \textbf{0.863821} &  0.106477 \\
infomap                      &  0.478261 &  0.713565 &  0.720074 &  0.153453 \\
infomap-weighted             &  0.433657 &  0.792426 &  0.748072 &  0.170349 \\
label\_propagation            &  0.364309 &  0.734100 &  0.717083 &  0.175682 \\
label\_propagation-weighted   &  0.478964 &  0.815100 &  0.750085 &  0.174347 \\
leading\_eigenvector          &  0.630606 &  0.783227 &  0.803901 &  0.116216 \\
leading\_eigenvector-weighted &  0.630435 &  0.846506 &  0.834201 &  0.107420 \\
walktrap                     &  0.667500 &  0.742232 &  0.781920 &  0.102776 \\
walktrap-weighted            &  \textbf{0.695652} &  0.856861 &  0.852037 &  0.091232 \\
\bottomrule
\end{tabular}

\end{adjustbox}
\label{tab:f1-micro.agg}
\caption{Likelihood of performing better than RakelD in Micro-averaged F1 score aggregated over data sets}
\end{table}

\begin{table}[H]
    \centering
    \begin{adjustbox}{width=1\textwidth}
    \small

\begin{tabular}{lrrrr}
\toprule
{} &   Minimum &    Median &      Mean &       Std \\
\midrule
BR                           &  0.500000 &  \textbf{0.985683} &  \textbf{0.919245} &  0.160273 \\
LP                           &  0.543478 &  0.829283 &  0.779482 &  0.163611 \\
fastgreedy                   &  0.543478 &  0.883312 &  0.866478 &  0.139572 \\
fastgreedy-weighted          &  \textbf{0.695652} &  0.969111 &  0.900483 &  0.116022 \\
infomap                      &  0.478261 &  0.820006 &  0.793086 &  0.147108 \\
infomap-weighted             &  0.500000 &  0.889556 &  0.818476 &  0.164492 \\
label\_propagation            &  0.449376 &  0.801890 &  0.754205 &  0.182183 \\
label\_propagation-weighted   &  0.521739 &  0.855195 &  0.821663 &  0.148561 \\
leading\_eigenvector          &  \textbf{0.695652} &  0.863565 &  0.851696 &  0.108860 \\
leading\_eigenvector-weighted &  \textbf{0.695652} &  0.889778 &  0.872119 &  0.118911 \\
walktrap                     &  \textbf{0.695652} &  0.846889 &  0.844807 &  0.105977 \\
walktrap-weighted            &  \textbf{0.695652} &  0.894335 &  0.885253 &  0.095436 \\
\bottomrule
\end{tabular}

\end{adjustbox}
\label{tab:f1-macro.agg}
\caption{Likelihood of performing better than RakelD in Macro-averaged F1 score aggregated over data sets}
\end{table}

\begin{table}[ht]
    \centering
    \begin{adjustbox}{width=1\textwidth}
    \small

\begin{tabular}{lrrrr}
\toprule
{} &   Minimum &    Median &      Mean &       Std \\
\midrule
BR                           &  0.000000 &  0.498000 &  0.558057 &  0.323240 \\
LP                           &  0.336570 &  0.958889 &  0.885476 &  0.190809 \\
fastgreedy                   &  0.213130 &  0.827043 &  0.791811 &  0.214756 \\
fastgreedy-weighted          &  0.061951 &  0.843413 &  0.747624 &  0.288200 \\
infomap                      &  \textbf{0.580500} &  \textbf{0.968667} &  \textbf{0.910039} &  0.143954 \\
infomap-weighted             &  0.525659 &  0.899430 &  0.859231 &  0.152637 \\
label\_propagation            &  0.380028 &  0.964444 &  0.900555 &  0.174854 \\
label\_propagation-weighted   &  0.336570 &  0.913652 &  0.861642 &  0.189690 \\
leading\_eigenvector          &  0.000000 &  0.826087 &  0.734935 &  0.340537 \\
leading\_eigenvector-weighted &  0.000000 &  0.843778 &  0.712555 &  0.345277 \\
walktrap                     &  0.078132 &  0.834338 &  0.739359 &  0.288072 \\
walktrap-weighted            &  0.429958 &  0.812667 &  0.810542 &  0.175723 \\
\bottomrule
\end{tabular}

\end{adjustbox}
\label{tab:accuracy.agg}
\caption{Likelihood of performing better than RakelD in Subset Accuracy aggregated over data sets}
\end{table}

\begin{table}[H]
    \centering
    \begin{adjustbox}{width=1\textwidth}
    \small

\begin{tabular}{lrrrr}
\toprule
{} &   Minimum &    Median &      Mean &       Std \\
\midrule
BR                           &  0.000000 &  0.369565 &  0.408252 &  0.375954 \\
LP                           &  0.000000 &  0.434653 &  0.380021 &  0.338184 \\
fastgreedy                   &  0.022222 &  0.469130 &  0.507034 &  0.266736 \\
fastgreedy-weighted          &  \textbf{0.093333} &  \textbf{0.554208} &  \textbf{0.558218} &  0.280209 \\
infomap                      &  0.000000 &  0.306667 &  0.396299 &  0.352963 \\
infomap-weighted             &  0.000000 &  0.332902 &  0.397576 &  0.345339 \\
label\_propagation            &  0.000000 &  0.457130 &  0.415966 &  0.361819 \\
label\_propagation-weighted   &  0.000000 &  0.489511 &  0.441813 &  0.363469 \\
leading\_eigenvector          &  0.044383 &  0.465511 &  0.456441 &  0.330251 \\
leading\_eigenvector-weighted &  0.000444 &  0.451556 &  0.457361 &  0.328774 \\
walktrap                     &  0.070667 &  0.438943 &  0.444424 &  0.312845 \\
walktrap-weighted            &  0.089333 &  0.379150 &  0.484974 &  0.331186 \\
\bottomrule
\end{tabular}

\end{adjustbox}
\label{tab:hl.agg}
\caption{Likelihood of performing better than RakelD in Hamming Loss aggregated over data sets}
\end{table}

\begin{table}[H]
    \centering
    \begin{adjustbox}{width=1\textwidth}
    \small

\begin{tabular}{lrrrr}
\toprule
{} &   Minimum &    Median &      Mean &       Std \\
\midrule
BR                           &  0.456522 &  0.778060 &  0.759178 &  0.202349 \\
LP                           &  0.355987 &  0.902632 &  0.854545 &  0.189086 \\
fastgreedy                   &  0.542302 &  0.876667 &  0.837570 &  0.135707 \\
fastgreedy-weighted          &  0.298197 &  0.875222 &  0.792386 &  0.205646 \\
infomap                      &  \textbf{0.650000} &  \textbf{0.945261} &  \textbf{0.889663} &  0.125510 \\
infomap-weighted             &  0.510865 &  0.889488 &  0.855242 &  0.148578 \\
label\_propagation            &  0.345816 &  0.928789 &  0.878622 &  0.181165 \\
label\_propagation-weighted   &  0.440592 &  0.920261 &  0.853821 &  0.181186 \\
leading\_eigenvector          &  0.163199 &  0.889874 &  0.821436 &  0.222219 \\
leading\_eigenvector-weighted &  0.464170 &  0.812444 &  0.794091 &  0.154102 \\
walktrap                     &  0.238558 &  0.891304 &  0.804866 &  0.227916 \\
walktrap-weighted            &  0.644889 &  0.863722 &  0.852369 &  0.122837 \\
\bottomrule
\end{tabular}

\end{adjustbox}
\label{tab:jaccard.agg}
\caption{Likelihood of performing better than RakelD in Jaccard Similarity aggregated over data sets}
\end{table}

\begin{table}[H]
    \centering

\begin{tabular}{lc}
\toprule
 & Iman-Davenport p-value \\
\midrule
accuracy & {\bf 0.0000000004} \\
f1-macro & {\bf 0.0000000000} \\
f1-micro & {\bf 0.0000000177} \\
hl & {\bf 0.0491215784} \\
jaccard & {\bf 0.0000124790} \\
\bottomrule
\end{tabular}
\label{tab:imans}
\caption{The p-values of the assessment of performance of the multi-label learning approaches compared against random baseline by the Iman-Davenport-Friedman multiple comparison, per measure}
\end{table}

\begin{table}[H]
    \centering
    \begin{adjustbox}{width=1\textwidth}
    \small

\begin{tabular}{llllll}
\toprule
 & accuracy & f1-macro & f1-micro & hl & jaccard \\
\midrule
BR & 0.3590121 & {\bf 0.0000003} & {\bf 0.0000500} & 1.0000000 & {\bf 0.0064229} \\
fastgreedy & 0.0515844 & {\bf 0.0000234} & {\bf 0.0000656} & 1.0000000 & {\bf 0.0038623} \\
fastgreedy.weighted & 0.1089704 & {\bf 0.0000001} & {\bf 0.0000001} & 0.3472374 & {\bf 0.0085647} \\
infomap & {\bf 0.0000257} & {\bf 0.0484159} & {\bf 0.0205862} & 1.0000000 & {\bf 0.0000705} \\
infomap.weighted & {\bf 0.0002717} & {\bf 0.0010778} & {\bf 0.0024092} & 1.0000000 & {\bf 0.0001198} \\
label.propagation & {\bf 0.0000112} & {\bf 0.0484159} & {\bf 0.0205862} & 1.0000000 & {\bf 0.0000098} \\
label.propagation.weighted & {\bf 0.0005315} & {\bf 0.0015858} & {\bf 0.0024092} & 1.0000000 & {\bf 0.0001196} \\
leading.eigenvector & 0.1860319 & {\bf 0.0001282} & {\bf 0.0002372} & 1.0000000 & {\bf 0.0056673} \\
leading.eigenvector.weighted & 0.2570154 & {\bf 0.0000274} & {\bf 0.0000653} & 1.0000000 & {\bf 0.0259068} \\
LP & {\bf 0.0000641} & {\bf 0.0280673} & {\bf 0.0205862} & 1.0000000 & {\bf 0.0000705} \\
walktrap & 0.0780264 & {\bf 0.0000397} & {\bf 0.0004457} & 1.0000000 & {\bf 0.0056673} \\
walktrap.weighted & {\bf 0.0192676} & {\bf 0.0000239} & {\bf 0.0000040} & 1.0000000 & {\bf 0.0018482} \\
\bottomrule
\end{tabular}

\end{adjustbox}
\label{tab:statistics}
\caption{The post-hoc pairwise comparison p-values of the assessment of performance of the multi-label learning approaches compared against random baseline by the Iman-Davenport-Friedman test with Rom post-hoc procedure, per measure}
\end{table}

\newpage

\bibliography{biblio}
\bibliographystyle{mdpi}

%

\end{document}